\documentclass[12pt, draftclsnofoot, onecolumn]{IEEEtran}
\usepackage{cite}
\usepackage{amsmath, amssymb, amsfonts, amsthm}
\usepackage{algorithm}
\usepackage[noend]{algorithmic}
\usepackage{graphicx}
\usepackage{textcomp}
\usepackage{xcolor}
\usepackage{color}
\usepackage{bm}
\usepackage{booktabs}
\usepackage{autobreak}
\usepackage{multirow}
\usepackage{booktabs}
\usepackage{authblk}
\usepackage{mathrsfs}
\usepackage{array}
\usepackage{makecell} 
\usepackage{float}
\usepackage{soul}
\usepackage{stmaryrd}
\usepackage{dsfont}
\usepackage{bbm}
\usepackage[hidelinks]{hyperref}
\usepackage{cleveref}
\usepackage{pifont}
\usepackage{enumitem}

\usepackage{subcaption}
\usepackage{placeins}

\newtheorem{theorem}{Theorem}

\newtheorem{corollary}{Corollary}

\begin{document}

\bstctlcite{BSTcontrol}

\title{Attention-Based Feature Online Conformal Prediction for Time Series}

\author{{Meiyi Zhu, Caili Guo \IEEEmembership{Senior Member, IEEE}, Chunyan Feng \IEEEmembership{Senior Member, IEEE}, and Osvaldo Simeone \IEEEmembership{Fellow, IEEE}
}


\thanks{{The work of M. Zhu and O. Simeone was supported by the Open Fellowships of the EPSRC (EP/W024101/1). The work of O. Simeone was also supported by the EPSRC project (EP/X011852/1).} The work of C. Guo and C. Feng was supported by the National Natural Science Foundation of China (61871047, 62371070) and by the Beijing Natural Science Foundation (L222043).

Meiyi Zhu and Osvaldo Simeone are with the King's Communications, Learning \& Information Processing (KCLIP) lab, Department of Engineering, King's College London, London WC2R 2LS, U.K. (e-mail: meiyi.1.zhu@kcl.ac.uk; osvaldo.simeone@kcl.ac.uk).

Caili Guo and Chunyan Feng are with the Beijing Key Laboratory of Network System Architecture and Convergence, School of Information and Communication Engineering, Beijing University of Posts and Telecommunications, Beijing 100876, China (e-mail: guocaili@bupt.edu.cn; cyfeng@bupt.edu.cn).

}
}

\maketitle

\begin{abstract}
Online conformal prediction (OCP) wraps around any pre-trained predictor to produce prediction sets with coverage guarantees that hold irrespective of temporal dependencies or distribution shifts. However, standard OCP faces two key limitations: it operates in the output space using simple nonconformity (NC) scores, and it treats all historical observations uniformly when estimating quantiles. This paper introduces attention-based feature OCP (AFOCP), which addresses both limitations through two key innovations. First, AFOCP operates in the feature space of pre-trained neural networks, leveraging learned representations to construct more compact prediction sets by concentrating on task-relevant information while suppressing nuisance variation. Second, AFOCP incorporates an attention mechanism that adaptively weights historical observations based on their relevance to the current test point, effectively handling non-stationarity and distribution shifts. We provide theoretical guarantees showing that AFOCP maintains long-term coverage while provably achieving smaller prediction intervals than standard OCP under mild regularity conditions. Extensive experiments on synthetic and real-world time series datasets demonstrate that AFOCP consistently reduces the size of prediction intervals by as much as $88\%$ as compared to OCP, while maintaining target coverage levels, validating the benefits of both feature-space calibration and attention-based adaptive weighting.
\end{abstract}

\section{Introduction}

\subsection{Context and Motivation}

Uncertainty quantification has become increasingly critical as machine learning systems are deployed in high-stakes applications such as autonomous vehicles, medical diagnosis, financial forecasting, and telecommunications  \cite{gawlikowski2021survey, angelopoulos2021gentle,simeone2025conformal}. While deep neural networks and language models have achieved remarkable predictive accuracy, they often produce overconfident predictions without reliable uncertainty estimates \cite{guo2017calibration,kadavath2022language}. Traditional approaches to uncertainty quantification, such as Bayesian methods \cite{gal2016dropout,simeone2022machine,huang2024calibrating} and ensemble techniques \cite{lakshminarayanan2017simple,abbasli2025comparing}, either require strong distributional assumptions or significant computational overhead, limiting their practical applicability.

Conformal prediction  (CP)  \cite{vovk2005algorithmic, shafer2008tutorial} offers an attractive alternative by providing distribution-free prediction sets with finite-sample validity guarantees. Given a target miscoverage rate $\alpha$, conformal prediction constructs prediction sets that contain the true label with probability at least $1-\alpha$, without making any assumptions about the underlying data distribution beyond exchangeability. This framework has gained significant attention in recent years due to its model-agnostic nature and rigorous theoretical foundations \cite{angelopoulos2024theoretical}.

However, the exchangeability assumption -- that data points can be reordered without changing their joint distribution -- is frequently violated in time series and sequential prediction tasks. Temporal dependencies, concept drift, and distribution shifts are inherent characteristics of many real-world applications, including financial markets, weather forecasting, and sensor networks. When exchangeability fails, standard CP methods can suffer from miscalibration, leading to prediction sets that either under-cover, failing to contain the true value, or over-cover, producing excessively large, uninformative intervals.

To address non-exchangeability, online conformal prediction (OCP) \cite{gibbs2021adaptive} adapts the conformal framework by continuously updating miscoverage levels, or prediction confidence thresholds,  through feedback mechanisms. Unlike CP, whose coverage guarantees are probabilistic, OCP provides deterministic long-term coverage guarantees. However, in practice, it still faces two important limitations:\begin{enumerate} \item \emph{Processing in the output space:} OCP typically leverages simple confidence scores in the output space, such as absolute prediction errors. This approach fails to leverage the rich semantic representations learned by modern deep neural networks, potentially leading to overly conservative prediction intervals. \item \emph{Uniform weighting:}  OCP treats all historical observations \emph{uniformly} when producing the current prediction set, ignoring the fact that some past data points may be more relevant than others for predicting uncertainty at the given time step. \end{enumerate}The goal of this work is to address these limitations.

\subsection{Related Work}

\subsubsection{Conformal Prediction}

Building on the theoretical foundations set in  \cite{vovk2005algorithmic}, the literature on CP has developed along several directions, including extensions to  inductive   \cite{papadopoulos2002inductive} and cross-validation-based methodologies \cite{lei2018distribution,cohen2024cross }. More recent  advances have extended conformal prediction to non-exchangeable settings. For example, reference \cite{barber2023conformal} developed weighted CP (WCP) methods with explicit bounds on coverage gaps under distribution drift, while the work \cite{oliveira2024split} proved that inductive CP remains approximately valid for non-exchangeable processes (see also \cite{tibshirani2019conformal,bhattacharyya2024group}). 

For time series, the works \cite{zecchin2024forking,lindemann2023safe} proposed methods that obtain coverage properties on average over time series, while the papers \cite{xu2021conformal,xu2023sequential} introduced methodologies that can  leverage existing predictors to achieve asymptotic valid conditional coverage under given technical assumptions. In contrast, references \cite{auer2023conformal} and \cite{chenconformalized} proposed methods that optimize the weights in WCP using attention-based mechanisms to identify similar regimes in time series. The coverage guarantees of these methods hinge of knowledge of the distribution shifts, which is practically unavailable.

The authors of \cite{teng2022predictive} introduced feature CP (FCP), which extends conventional CP  to operate in semantic feature spaces by leveraging the inductive bias of deep representation learning. They demonstrate provable improvements over output-space methods under reasonable assumptions, such as a stable feature space and a smooth prediction head.

\subsubsection{Online Conformal Prediction}
The reference \cite{gibbs2021adaptive} introduced OCP, which continuously re-estimates miscoverage parameters using gradient descent to maintain long-term coverage under distribution shift. Follow-up work extended OCP with automatic step-size tuning \cite{gibbs2024conformal,zaffran2022adaptive}, proportional-integral-derivative mechanisms \cite{angelopoulos2023conformal}, localization mechanisms \cite{zecchin2024localized}, and  strongly adaptive properties \cite{bhatnagar2023improved}.

\subsection{Main Contributions}

\begin{figure*}[t]
    \centering
    \includegraphics[width=0.63\textwidth]{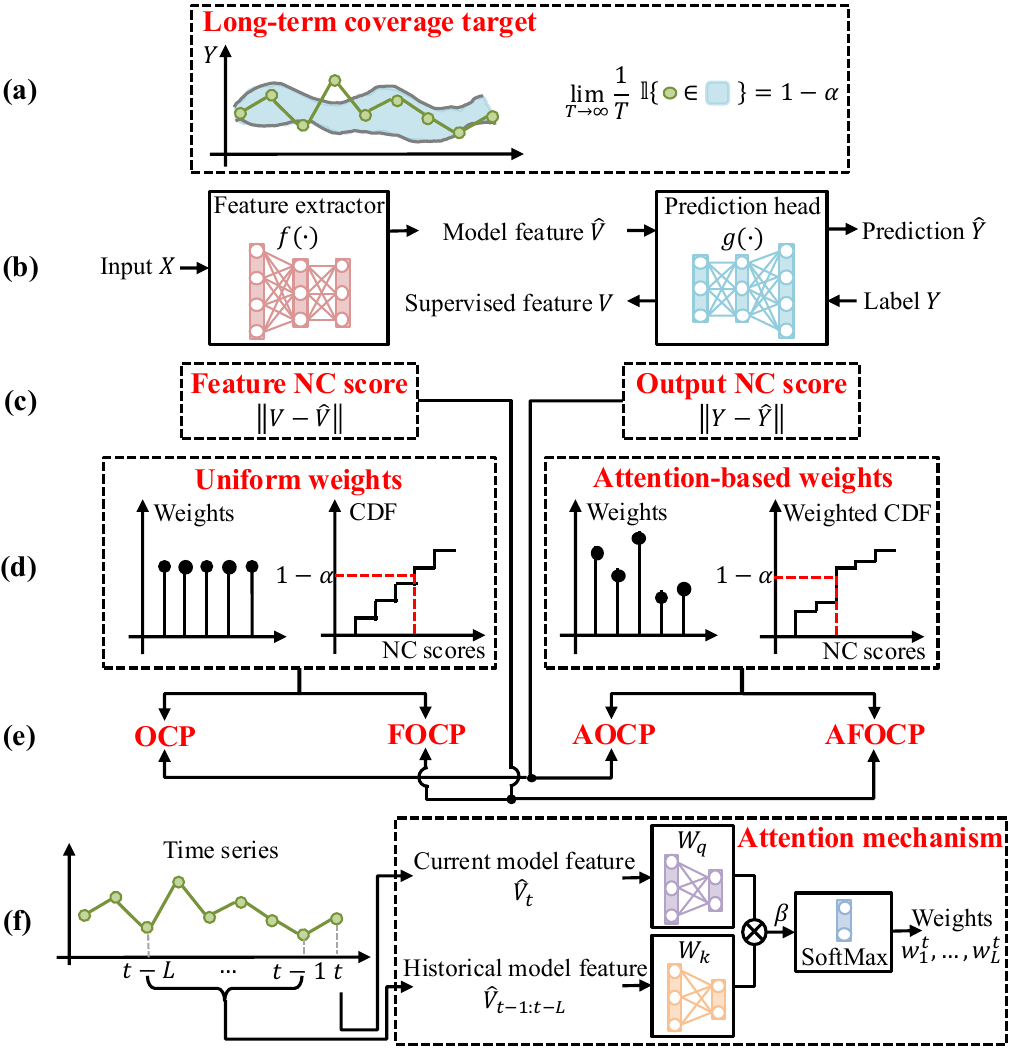}
    \vspace{-1mm}
    \caption{Overview of AFOCP and related baselines. (a) The goal of this work is to calibrate pre-trained predictors by augmenting their outputs with prediction sets that contain the true label $Y$ for a fraction at least $100(1-\alpha)\%$ of the time. (b) For any input $X$, the pre-trained model $\mu (X) = g \circ f(X)$ maps inputs $X$ through the feature extractor $f(\cdot)$ and the prediction head $g(\cdot)$. (c) Nonconformity (NC) scores can be evaluated in the \textit{output} or \textit{feature} spaces. (d) The NC scores can be combined to evaluate empirical distributions and quantiles using either \textit{uniform weights} or \textit{attention-based weights}. (e) OCP \cite{gibbs2021adaptive} uses output scores with uniform weights; feature OCP (FOCP) uses feature scores, while retaining uniform weights; attention-based OCP (AOCP) keeps output scores but learns data-dependent weights via attention; and attention-based feature OCP (AFOCP) combines feature scores with attention-based weights. FOCP, AOCP, and AFOCP are introduced in this work, with AFOCP being the most general of the three. (f) The attention mechanism in AOCP and AFOCP compares the current model feature with past features to produce similarity-based weights that serve as data-dependent weights for calibration.}
    \label{fig_sys_mod}
    \vspace{-8mm}
\end{figure*}

In this paper, we aim to calibrate a pre-trained machine learning model to generate a prediction set that includes the ground-truth label for a sufficiently large fraction of time. Specifically, as shown in Fig. \ref{fig_sys_mod}, we consider an online setting in which input-label pairs $\left\{\left(X_t, Y_t\right)\right\}_{t=1,2,\ldots}$ arrive sequentially as a time series over discrete time $t$. At time $t=1,2,\ldots$, given a test input $X_t$, a pre-trained model $\mu(\cdot)$, and the historical data pairs $\left\{\left(X_{\tau}, Y_{\tau} \right) \right\}_{\tau=1}^{t-1}$, our goal is to construct a prediction set that contains the true label $Y_t$ for a fraction at least $100(1-\alpha)\%$ of the time.

As reviewed above, this goal can be attained by leveraging OCP and variants. This paper introduces \emph{attention-based feature OCP} (AFOCP), a novel framework that  enhances OCP through two key innovations:\begin{enumerate}
\item \emph{Feature-space OCP for compact prediction sets:} We extend OCP to construct prediction sets in the feature space of pre-trained neural networks. By computing confidence scores using learned representations rather than output-space predictions, AFOCP exploits the inductive bias of deep learning models to construct more informative prediction sets. Specifically, the feature extractor allows AFOCP to concentrate on task-relevant information, while suppressing nuisance variation. 

\item \emph{Attention-based adaptive weighting for non-stationary data:} We incorporate an attention mechanism that learns to assign relevance weights to historical observations based on their similarity to the current test point in feature space. Unlike standard OCP, which treats all data in the calibration window uniformly, our attention-based approach emphasizes past observations from similar distributional regimes. The attention weights are learned online through an autoregressive prediction task, where the model minimizes the error in predicting current nonconformity (NC) scores from past scores, weighted by feature similarity. This enables AFOCP to adapt to distribution shifts and temporal dependencies without requiring explicit change-point detection or regime identification.
\end{enumerate}
Overall, our main contributions are as follows:
\begin{itemize}
\item We propose FOCP, an online calibration approach operating in the learned feature space of a pre-trained predictor. We further generalize FOCP to AFOCP by replacing uniform score aggregation with attention-based weighting over recent observations.

\item We establish deterministic long-term coverage guarantees in the online setting and show that FOCP and AFOCP can attain strictly shorter time-averaged prediction interval lengths than output-space OCP, or its attention-based counterpart, AOCP, under some regularity assumptions.
\item On synthetic and real-world time-series benchmarks, AFOCP consistently achieves the target coverage, while markedly reducing prediction interval length relative to OCP. Ablations are provided to disentangle the roles of feature-space calibration and attention-based weighting, showing how their relative gains vary with the feature dimension and calibration window length.
\end{itemize}

The remainder of this paper is organized as follows. Section \ref{sec_on_CP} reviews OCP. Section \ref{sec_on_AFCP} introduces the AFOCP framework, including feature-based NC scores, attention-based weighting, and theoretical guarantees. Section \ref{sec_experiment} presents experimental results on synthetic and real-world datasets. Section \ref{sec_conclusion} concludes with discussion and future directions.

\section{Online Conformal Prediction} \label{sec_on_CP}
OCP \cite{gibbs2021adaptive} extends the traditional CP framework \cite{vovk2005algorithmic} by incorporating an online update mechanism. In this setting, CP defines an NC score to measure the dissimilarity between the model's prediction $\mu(X)$ for an input $X\in\mathcal{X}\subseteq \mathbb{R}^{D_{\textrm{in}}}$ and any candidate label $Y\in\mathcal{Y}\subseteq\mathbb{R}^{D_{\textrm{out}}}$, where $D_{\textrm{in}}$ and $D_{\textrm{out}}$ denote the dimensions of input and output variables, respectively. For regression tasks, a common choice for the NC score is the absolute error, i.e.,
\begin{align}\label{eq_NC}
    s(X, Y) = \left\|Y-\mu(X)\right\|.
\end{align}

To adapt the method for streaming data, at time $t$, we calculate the $(1-\alpha_t)$-quantile of the NC scores using a sliding window of the most recent $L$ observations $\left\{(X_{t-\tau}, Y_{t-\tau}) \right\}_{\tau=1}^L$. The prediction set for input $X_t$ is then constructed as
\begin{align}\label{eq_CP_pred_set}
    \Gamma^{\textrm{OCP}}_t(X_t) = \left\{Y\in\mathcal{Y}:s\left(X_t, Y\right) \leq Q_{1-\alpha_t} \left(\sum_{\tau=1}^{L} \frac{1}{L+1} \delta_{s(X_{t-\tau}, Y_{t-\tau})} + \frac{1}{L+1}\delta_{+\infty}\right)\right\},
\end{align}
where $Q_a(\cdot)$ denotes the $a$-quantile of its argument and $\delta_b$ represents the Dirac delta function centered at $b$. We assume that $\alpha_1\in[0,1]$ and that the quantile function $Q_a$ is non-decreasing, with $Q_a=-\infty$ for $a<0$ and $Q_a=\infty$ for $a>1$.

For each prediction set $\Gamma^{\textrm{OCP}}_t(X_t)$ to achieve coverage probability $1-\alpha$, i.e., $\mathbb{P} \left(Y_t\in\Gamma^{\textrm{OCP}}_t(X_t)\right) \geq 1-\alpha$, the data $\left\{\left(X_{\tau}, Y_{\tau}\right)\right\}_{\tau=1}^t$ must be exchangeable, and we assume a fixed $\alpha_t=\alpha$ \cite{vovk2005algorithmic}. However, this exchangeability assumption typically does not hold for time-series data. To address this, OCP further updates the unreliability level $\alpha_t$ in the quantile in \eqref{eq_CP_pred_set} using the online rule \cite{gibbs2021adaptive},
\begin{align}\label{eq_update1}
    \alpha_{t+1}=\alpha_t + \gamma(\alpha-\textrm{err}_t),
\end{align}
where $\gamma$ denotes the step size, and the discrete error $\textrm{err}_t$ is defined as
\begin{align}
    \textrm{err}_t \triangleq
    \begin{cases}
       1 & \text{if $Y_t\notin \Gamma^{\textrm{OCP}}_t(X_t)$}, \\
       0 & \text{otherwise}.
    \end{cases}
\end{align}

This online update rule ensures that the predicted coverage probability converges to the desired level in the long run. The following theorem provides a reliability guarantee for OCP.
\begin{theorem}[\textbf{Proposition 4.1 \cite{gibbs2021adaptive}}]\label{theo_ACI}
    The average error over time $T\in\mathbb{N}$ is upper bounded as
    \begin{align}\label{theo_eq_err_up}
        \frac{1}{T} \sum_{t=1}^T\text{\rm{err}}_t \leq \alpha + \frac{\alpha_1-\alpha_{T+1}}{T\gamma}.
    \end{align}
    In particular, as $T\rightarrow\infty$, the error converges to the desired level $\alpha$, i.e.,
    \begin{align}\label{theo_eq_err_gua}
        \lim_{T\rightarrow\infty} \frac{1}{T} \sum_{t=1}^T \text{\rm{err}}_t = \alpha.
    \end{align}
\end{theorem}
\textit{Proof:}
Expanding the recursion in \eqref{eq_update1} leads to \eqref{theo_eq_err_up}. Since $\alpha_t\in[-\gamma, 1+\gamma]$ according to Lemma 4.1 in \cite{gibbs2021adaptive}, the long-term coverage in \eqref{theo_eq_err_gua} holds.

\section{Attention-based Feature Online Conformal Prediction}\label{sec_on_AFCP}
In this paper, we introduce AFOCP, a novel variant of OCP that leverages compact data representations in the feature space in order to obtain more compact prediction sets and to learn relevance weights over historical data to inform an attention-based quantile estimate.

\subsection{Feature Online Conformal Prediction}\label{sec_FOCP}
As done in \cite{teng2022predictive} to introduce feature (offline) CP, we focus on pre-trained prediction models $\mu(\cdot)$ that can be decomposed as
\begin{align}\label{eq_model}
    \mu(\cdot) = g\circ f(\cdot),
\end{align}
where the feature extractor $f(\cdot)$ maps the input $X$ to the latent feature $\hat{V}=f(X)\in\mathbb{R}^{D}$ with $D$ denoting the dimension of the feature, and the prediction head $g(\cdot)$ transforms these features into output predictions $\hat{Y}=g(\hat{V})$.

To any data pair $(X,Y)\in\mathcal{X}\times \mathcal{Y}$, we can thus associate two, generally different, features:
\begin{itemize}
    \item [1)] Model feature: The model feature is obtained by running the model in inference mode through the feature extractor $f(\cdot)$ as
    \begin{align}\label{eq_feature}
        \hat{V}=f(X)
    \end{align}
    \item [2)] Supervised feature: The supervised feature is obtained by running the model backward, starting from the label $Y$ through the inverse of the prediction head $g(\cdot)$ as
    \begin{align}\label{eq_su_feature}
        V\in g^{-1}\left(Y\right).
    \end{align}
    Since function $g(\cdot)$ is generally many-to-one, any vector $V$ in the inverse image $g^{-1}(Y)$ can be selected in \eqref{eq_su_feature}.
\end{itemize}

The NC score function in the feature space is then defined as the difference between model feature and supervised feature as \cite{teng2022predictive}
\begin{align}\label{eq_fcp_NC}
    s^{\textrm{f}}(X,Y) = \inf_{V\in g^{-1}\left(Y\right)}\left\|V-\hat{V}\right\|,
\end{align}
where the infimum operator is over all feature vectors in the inverse image \eqref{eq_su_feature} of the prediction head. We discuss in Sec. \ref{subsec_eval_NC} how to evaluate the scores \eqref{eq_fcp_NC} in practice.

Given a new test input $X_t$ at time $t$, the proposed AFOCP scheme constructs a prediction set $\Gamma^{\textrm{AFOCP}}_t(X_t)$ based on scores in the feature space for the latest observed $L$ data pairs $\left\{\left(X_{t-\tau}, Y_{t-\tau}\right)\right\}_{\tau = 1}^L$. Specifically, the prediction set for input $X_t$ is constructed as
\begin{align}\label{eq_fCP_pred_set}
    \Gamma^{\textrm{AFOCP}}_t(X_t) = \left\{Y\in\mathcal{Y}:s^{\textrm{f}}\left(X_t,Y\right)\leq Q_{1-\alpha_t}\left(\sum_{\tau=1}^{L} w_t^{\tau} \delta_{s^{\textrm{f}}\left(X_{t-\tau}, Y_{t-\tau}\right)} + w_t^{L+1} \delta_{+\infty}\right)\right\},
\end{align}
where the quantile function $Q_a(\cdot)$ and point mass $\delta_b$ are defined as in \eqref{eq_CP_pred_set}, while the weights $\left\{w_t^{\tau}\right\}_{\tau=1}^{L+1}$ are discussed next. Compared to the conventional OCP predictor in \eqref{eq_CP_pred_set}, AFOCP has the following distinguishing characteristics:
\begin{itemize}
    \item [1)] It leverages feature-based NC scores to evaluate the prediction set \eqref{eq_fCP_pred_set}. This typically yields shorter prediction sets because the learned representation $f(X)$ concentrates task-relevant information and suppresses nuisance variation, which reduces the dispersion of feature-based NC scores \cite{teng2022predictive}. A smooth prediction head $g$ maps this reduction to the output, while preserving the target coverage. See Theorem \ref{theo_on_fcp_band} for a rigorous justification.

    \item [2)] It introduces normalized weights $\left\{w_t^{\tau}\right\}_{\tau=1}^{L+1}$, with weight $w_t^{\tau}$ assigned to the data point $X_{t-\tau}$ for $\tau=1, \ldots, L$. Ideally, a larger weight $w^{\tau}_t$ should be assigned to a data point $X_{t-\tau}$ that is likely to share a similar data distribution with the current test point $X_t$, thereby improving the accuracy of the quantile estimation \cite{tibshirani2019conformal, barber2023conformal, auer2023conformal, chenconformalized}. The weights $\left\{w^\tau_t\right\}_{\tau=1}^{L+1}$ satisfy the normalization condition $\sum_{\tau=1}^{L+1} w^{\tau}_t=1$ with each $w^{\tau}_t\in[0,1]$. AFOCP applies a weight assignment strategy that follows an attention mechanism detailed in Sec. \ref{sec_on_att}.
\end{itemize}

To maintain the long-term reliability, the unreliability level $\alpha_t$ in \eqref{eq_fCP_pred_set} is updated via an online update rule similar to \eqref{eq_update1}, i.e.,
\begin{align}\label{eq_unrelia_lvl}
    \alpha_{t+1} = \alpha_t + \lambda(\alpha-\textrm{err}^\textrm{f}_t),
\end{align}
where $\lambda$ denotes the step size and the discrete error $\textrm{err}^{\textrm{f}}_t$ is defined as
\begin{align}
    \textrm{err}^{\textrm{f}}_t \triangleq
    \begin{cases}
       1 & \text{if $Y_t\notin \Gamma^{\textrm{AFOCP}}_t(X_t)$}, \\
       0 & \text{otherwise}.
    \end{cases}
\end{align}

\subsection{Online Update of Weights via Attention Mechanism} \label{sec_on_att}
In AFOCP, the weights $\left\{w_t^\tau\right\}_{\tau=1}^{L+1}$ in \eqref{eq_fCP_pred_set} are obtained via an attention mechanism that aims at capturing the feature-space similarity between the current feature vector $f(X_t)$ and the feature vectors of the $L$ most recent observations, i.e.,
\begin{align}\label{eq_feat_L}
    \hat{V}_{t-1:t-L} \triangleq \left[f(X_{t-1}), \ldots, f(X_{t-L})\right]^{\mathrm{T}}.
\end{align}

Accordingly, first we compute the attention vector
\begin{align}\label{eq_attn_wei}
    \left[a_t^1, \ldots, a_t^L \right]^{\mathrm{T}} \triangleq \textrm{attention}\big(f(X_t), V_{t-1: t-L}\big)\in\mathbb{R}^{1\times L},
\end{align}
where each attention coefficient $a_t^l$ quantifies the similarity of the feature vectors $f(X_t)$ and $f(X_{t-l})$. {Specifically, given two sequences $\left\{U_1, \ldots, U_{L_q}\right\}$ and $\left\{V_1, \ldots, V_{L_k}\right\}$ with $U_i\in\mathbb{R}^D$ and $V_j\in\mathbb{R}^D$, the attention operator returns the $\mathbb{R}^{L_q\times L_k}$ matrix whose $(i,j)$-th entry is
\begin{align}\label{eq_attn_mech}
    \textrm{attention}\big(\left\{U_1, \ldots, U_{L_q}\right\}, \left\{V_1, \ldots, V_{L_k} \right\}\big)_{(i,j)} = \frac{\exp\big(\beta\langle U_i W_q,V_j W_k\rangle\big)}{\sum_{j'=1}^{L_k}\exp\big(\beta\langle U_iW_q, V_{j'}W_k\rangle\big)},
\end{align}
where $W_q\in\mathbb{R}^{D\times D'}$ and $W_k\in\mathbb{R}^{D\times D'}$ are the learned query and key embedding matrices with latent dimension $D'$, $\beta>0$ is a scaling factor, and $\langle \cdot, \cdot\rangle$ denotes the standard Euclidean inner product on $\mathbb{R}^{D'}$.}

The attention coefficients \eqref{eq_attn_wei} are then re-normalized to obtain the weights used in \eqref{eq_fCP_pred_set} as
\begin{subequations}\label{eq_att_wei}
    \begin{align}
        \left[w_t^1, \ldots, w_t^L\right]^{\mathrm{T}} &= \frac{L}{L+1} \cdot \left[a_t^1, \ldots, a_t^L \right]^{\mathrm{T}},\\
        \textrm{and} ~ w_t^{L+1} &= \frac{1}{L+1},
    \end{align}
\end{subequations}
where weight $w^{L+1}_t$ is assigned to the $+\infty$ point mass in \eqref{eq_fCP_pred_set}. A larger weight $w^\tau_t$ indicates greater relevance of the $\tau$-th latest historical feature $\hat{V}_{t-\tau}$ to $f(X_t)$.

The attention mechanism \eqref{eq_attn_mech} depends on the embedding matrices $W_q$ and $W_k$. We propose to optimize these matrices in an online fashion by addressing a linear prediction problem on the NC scores. To elaborate, for each time $t$, denote the NC scores evaluated on the most recent $L$ pairs $\left\{(X_{t-\tau},Y_{t-\tau})\right\}_{\tau=1}^{L}$ as
\begin{align}
    S^{\textrm{f}}_{t-1:t-L} \triangleq \left[s^{\textrm{f}}(X_{t-1},Y_{t-1}), \ldots, s^{\textrm{f}}(X_{t-L}, Y_{t-L})\right]^{\mathrm{T}}, ~ \textrm{with} ~ S^{\textrm{f}}_t \triangleq s^{\textrm{f}}(X_t, Y_t).
\end{align}
We adopt a linear predictor that aggregates the past $L$ scores to predict the current score using the attention weights \eqref{eq_attn_wei}, i.e.,
\begin{align}
    \hat{S}_t^\textrm{f} \triangleq \sum_{\tau=1}^L a_t^{\tau} \cdot S^{\textrm{f}}_{t - \tau}.
\end{align}
{We take the cumulative squared prediction error
\begin{align}\label{eq_att_loss}
    \mathcal{L}_t \triangleq \sum_{\tau=1}^t\big(S^{\textrm{f}}_{\tau} - \hat{S}^{\textrm{f}}_{\tau}\big)^2
\end{align}
as our optimization objective, and update $W_q$ and $W_k$ in \eqref{eq_attn_mech} online by gradient descent on the instantaneous loss $\big(S^{\textrm{f}}_{\tau} - \hat{S}^{\textrm{f}}_{\tau}\big)^2$. This encourages the attention mechanism \eqref{eq_attn_mech} to assign larger weights to the past indices whose scores best predict the current score $S^{\textrm{f}}_t$.
}

\subsection{Construction of the Prediction Set}\label{subsec_eval_NC}
A practical challenge in constructing the prediction set $\Gamma^{\textrm{AFOCP}}_t (X_t)$ in \eqref{eq_fCP_pred_set} lies in evaluating the feature-based NC score $s^\textrm{f}(X,Y)$ in \eqref{eq_fcp_NC} due to the need to evaluate the infimum operator. Following \cite{teng2022predictive}, we approximate the score $s^{\textrm{f}}(X,Y)$ using gradient descent in the feature space. Initializing the solution $V$ at the model feature $f(X)$ and using a step size $\eta>0$ yields the update
\begin{align}\label{eq_approx_sup_feat}
    \quad V \leftarrow V - \eta \nabla_V \|g(V) - Y\|^2.
\end{align}
{We stop the gradient descent after a fixed number $N$ of iterations, and denote the final iterate by $\bar{V}$.} The NC score is then approximated as $S^{\textrm{f}}\approx\|\bar{V} - f(X)\|$. Note that this is an upper bound on the true NC score $s^{\textrm{f}}(X,Y)$.

Another practical challenge is the composition of the set \eqref{eq_fCP_pred_set} given the most recent $L$ feature-based NC scores $\left\{s^\textrm{f}(X_{t-\tau},Y_{t-\tau})\right\}_{\tau=1}^L$. In fact, when the label space $\mathcal{Y}$ is discrete and finite, as in classification tasks, one can construct the set \eqref{eq_fCP_pred_set} by enumerating the possible labels $Y\in \mathcal{Y}$. In contrast, when $\mathcal{Y}$ is continuous, as in regression tasks, it is not generally feasible to calculate the NC scores $s^{\textrm{f}}\left(X_t,Y\right)$ across all candidates $Y\in\mathcal{Y}$. Following \cite{teng2022predictive}, we instead adopt a band estimation strategy based on linear relaxation based perturbation analysis (LiRPA) \cite{xu2020automatic}. This scheme provides a certified upper-bound approximation of prediction set \eqref{eq_fCP_pred_set} via linear relaxation. The resulting interval is then a computationally efficient outer approximation of $\Gamma_t^{\textrm{AFOCP}}\left(X_t\right)$. We refer to \cite{teng2022predictive} for details.

\begin{algorithm}[t]
  \caption{AFOCP}\label{algo_on_AFCP}
  \hspace*{\algorithmicindent}\parbox[t]{\dimexpr\linewidth-\algorithmicindent}{\textbf{Input:} Data stream $\left\{\left(X_t,Y_t\right)\right\}_{t\in\mathbb{N}}$; pre-trained machine learning model $\mu\left(\cdot\right) = g \circ f(\cdot)$; target long-term miscoverage $\alpha$; step sizes $\lambda$ and $\eta$; scaling factor $\beta$; number of gradient steps $N$}
  \begin{algorithmic}[1]
    \FOR{each time step $t$}
        \STATE{Compute weights $\left\{w_t^\tau\right\}_{\tau=1}^{L+1}$ via the attention mechanism \eqref{eq_attn_wei}}
        \STATE{Calculate the feature-based NC scores $\left\{s^{\textrm{f}}\left(X_{t-\tau},Y_{t-\tau}\right)\right\}_{\tau=1}^L$ using \eqref{eq_approx_sup_feat}}
        \STATE{Form the weighted empirical distribution in \eqref{eq_fCP_pred_set} and compute its $(1-\alpha_t)$-quantile}
        \STATE{Construct the prediction set $\Gamma_t^{\textrm{AFOCP}}\left(X_t\right)$ in \eqref{eq_fCP_pred_set} (see Sec. \ref{sec_FOCP})}
        \STATE{Update the miscoverage level $\alpha_t$ using \eqref{eq_unrelia_lvl}}
        \STATE{Train the attention mechanism $\textrm{attention}(\cdot,\cdot)$ online by minimizing the loss \eqref{eq_att_loss} over $W_q$ and $W_k$}
    \ENDFOR
  \end{algorithmic}
\end{algorithm}

\subsection{Special Cases of AFOCP}
In order to isolate and evaluate the effectiveness of feature-based NC scores and attention-based weights, we also introduce two intermediate variants of OCP, namely, AOCP and FOCP, as illustrated in Fig. \ref{fig_sys_mod}.

\subsubsection{AOCP} NC scores are computed in the output space as in OCP, but uniform weights are replaced by attention-based adaptive weights. The prediction set is defined as
\begin{align}\label{eq_AOCP}
    \Gamma^{\textrm{AOCP}}_t(X_t) = \left\{Y\in\mathcal{Y}:s\left(X_t,Y\right)\leq Q_{1-\alpha_t}\left(\sum_{\tau=1}^{L} w_t^{\tau} \delta_{s\left(X_{t-\tau}, Y_{t-\tau}\right)} + w_t^{L+1} \delta_{+\infty}\right)\right\}.
\end{align}
Unlike OCP, which uses uniform weights for the last $L$ NC scores in \eqref{eq_CP_pred_set}, AOCP assigns data-dependent weights via an attention mechanism. The weight generation follows the same procedure as AFOCP described in Sec. \ref{sec_on_att}, except that the NC scores in the loss function \eqref{eq_att_loss} are computed using \eqref{eq_NC} in the output space instead of the feature space.

\subsubsection{FOCP} The weighting scheme remains uniform as in OCP, but NC scores are evaluated in the semantic feature space. The prediction set is given by
\begin{align}\label{eq_FOCP}
    \Gamma^{\textrm{FOCP}}_t(X_t) = \left\{Y\in\mathcal{Y}:s^{\textrm{f}}\left(X_t,Y\right) \leq Q_{1-\alpha_t}\left(\sum_{\tau=1}^L \frac{1}{L+1}\delta_{s^{\textrm{f}}\left(X_{t-\tau}, Y_{t-\tau}\right)} + \frac{1}{L+1}\delta_{+\infty}\right)\right\}.
\end{align}
That is, FOCP replaces the output-level NC score calculation in \eqref{eq_NC} with the feature-level counterpart defined in \eqref{eq_fcp_NC}.

\subsection{Theoretical Guarantees} \label{sec_threo}
We now demonstrate that the proposed AFOCP scheme, summarized in Algorithm \ref{algo_on_AFCP}, ensures the desired long-term coverage, while being provably more efficient than the vanilla OCP reviewed in Sec. \ref{sec_on_CP}. First, similar to Theorem \ref{theo_ACI}, we have the following reliability guarantee.
\begin{corollary}\label{coro_on_fcp}
    The average error over $T\in\mathbb{N}$ obtained by FOCP and AFOCP can be upper bounded as
    \begin{align}
        \frac{1}{T} \sum_{t=1}^T\text{\rm{err}}^{\text{\rm{f}}}_t \leq \alpha + \frac{\alpha_1-\alpha_{T+1}}{T\lambda}.
    \end{align}
    In particular, we have the limit
    \begin{align}
        \lim_{T\rightarrow\infty} \frac{1}{T} \sum_{t=1}^T \text{\rm{err}}^{\text{\rm{f}}}_t = \alpha.
    \end{align}
\end{corollary}
\textit{Proof:}
The proof follows directly from Theorem \ref{theo_ACI}.

Second, to compare efficiency in terms of prediction interval length, we proceed under informal assumptions analogous to those underlying Theorem 4 in \cite{teng2022predictive}. While a formal statement can be found in Appendix \ref{sec_proof}, these assumptions essentially require that: (\emph{i}) on average over time, the output-space quantiles induced by the feature-space construction remain close to those based directly on output-space lengths; (\emph{ii}) the mapping from feature space to output space amplifies deviations between individual lengths and their quantiles;  and (\emph{iii}) the resulting output-space quantiles are temporally stable, with fluctuations diminishing as the calibration window grows. This yields the following result.

\begin{theorem}\label{theo_on_fcp_band}
    Under mild assumptions (see Appendix \ref{sec_proof}), for all time $T\in\mathbb{N}$, FOCP and AFOCP provably outperform OCP and AOCP, respectively, in terms of time-averaged prediction interval length, i.e.,
    \begin{align}
        \frac{1}{T}\sum_{t=1}^T\left|\Gamma^{\text{\rm{AFOCP/FOCP}}}_t(X_t)\right| \leq \frac{1}{T}\sum_{t=1}^T\left|\Gamma^{\text{\rm{AOCP/OCP}}}_t(X_t)\right|.
    \end{align}
\end{theorem}

Proof: Appendix \ref{sec_proof}.

\section{Numerical Results} \label{sec_experiment}
In this section, we empirically evaluate AFOCP and the baselines OCP, FOCP, and AOCP on one synthetic and four real-world time-series datasets. We report time-averaged coverage and time-averaged prediction interval length to verify the long-term coverage guarantees and to assess the efficiency of the resulting prediction sets. By varying the calibration window length and feature dimension, we also isolate the impact of attention-based adaptive weighting and feature-space calibration.

\subsection{Setting}
\subsubsection{Datasets}\label{sec_data_set}
Following the evaluation setting in \cite{chenconformalized}, we analyze the performance of OCP, FOCP, AOCP and AFOCP (see Sec. \ref{sec_on_AFCP}) on synthetic data and four real-world benchmark time-series datasets:
\begin{itemize}
    \item \textbf{Synthetic data \cite{auer2023conformal}:} We generate a multivariate time series of length $1500$ with alternating segments of variable lengths with different inputs and noise distributions. In particular, at each time step $t$, the input $X_t\in\mathbb{R}^{50}$ and target $Y_t\in\mathbb{R}^{50}$ are related as $Y_t = 10 + W X_t + \varepsilon_t$, where $W\in\mathbb{R}^{50\times 50}$ has i.i.d. entries drawn from $\mathcal{N}(0, 1/50)$. Each segment length is drawn uniformly from the set $\left\{40, 41, 42, \ldots, 80\right\}$. In a segment, the input $X_t$ is a vector of all entries equal to $3$ and the noise is $\varepsilon_t\sim \mathcal{N}(0, X_t/2 \mathbf{I}_{50})$; in the next segment, the next input $X_t$ is a vector of all entries equal to $21$ and the noise $\varepsilon_t$ has i.i.d. entries sampled from $\mathcal{U}(-X_t, X_t)$.

    \item \textbf{Air quality \cite{zhang2017cautionary, auer2023conformal}:} This dataset provides hourly air quality and meteorological measurements from the Tiantan station in Beijing from March $2013$ to February $2017$, totaling $35064$ timestamps. At each time step $t$, we predict the current particulate-matter concentration from an $11$-dimensional input $X_t$ built from other pollutant indicators, temperature, pressure, dew point, precipitation, wind speed, and wind direction encoded as two orthogonal components scaled to $[-1,1]$. The scalar target $Y_t$ alternates by contiguous segments between PM10 and PM2.5, with segment lengths uniformly sampled from the set $\left\{40, 41, 42, \ldots,80\right\}$, modeling a single sensor that intermittently measures the two particulate indicators in separate intervals.

    \item \textbf{Electricity \cite{harries1999splice, barber2023conformal}:} This dataset comprises half-hourly records of electricity price, demand, and transfer for New South Wales and Victoria from $7$ May $1996$ to December $1998$. Inputs $X_t$ are given by the state-level electricity prices and demands. The prediction target $Y_t$ is transfer, defined as the amount of electricity exchanged between the two states. We retain observations in the time slot $09\colon \hspace{-1mm}00–12\colon \hspace{-1mm}00$ and discard an initial transient period with constant transfer, yielding $3444$ time points.

    \item \textbf{Bike-sharing \cite{teng2022predictive}:} This dataset contains daily records from an urban bike-sharing system from $2011$ to $2012$, comprising $731$ days. At each time step $t$, the input $X_t$ has $16$ features capturing calendar context, such as year, month, weekday, holiday status, working day status, as well as weather conditions, such as air temperature, perceived temperature, humidity, wind speed, and a coarse weather category. The target $Y_t$ is the total number of rentals per day.

    \item \textbf{Wind speed \cite{xu2021conformal, dong2021multi}:} This dataset contains wind speed measurements from wind farms operated by the Midcontinent Independent System Operator in the United States, sampled every $15$ minutes over one week in September $2020$ and comprising $764$ timestamps. At each time step $t$, we use ten input features $X_t$ and a two-component target $Y_t$, with the inputs capturing contemporaneous values and short-term lags, and the targets describing future wind conditions.
\end{itemize}

For datasets exceeding $2000$ samples, we deterministically downsample to $2000$ evenly spaced observations while preserving temporal order, keeping the sequential online evaluation computationally tractable. We use an $85\%/15\%$ train/test split and report results averaged over five random seeds for robustness.

\subsubsection{Model training}
We train a two-stage neural network \eqref{eq_model} for regression, where both the feature extractor $f(\cdot)$ and the prediction head $g(\cdot)$ are two-layer fully connected networks with ReLU activations and hidden size $D$. The model is trained using mean squared error (MSE) loss, optimized by the Adam optimizer with a learning rate of $5\times 10^{-4}$, a weight decay of $10^{-6}$, and a batch size of $64$ for $10$ epochs.

We further pre-train an attention module to assign dynamic weights to historical data based on feature similarity. The input embeddings, produced by the feature extractor $f(\cdot)$, have dimension $D$, and are mapped into query and key vectors via linear layers with hidden dimension $D' = 32$, as in \eqref{eq_attn_mech}.
The module is first pre-trained on the training set using a sliding window of length $L$, optimizing the loss in \eqref{eq_att_loss} with the Adam optimizer (learning rate $5\times 10^{-4}$, weight decay $10^{-6}$) for $20$ epochs.
In the online phase, after each prediction, the new data point is appended to the window, which is then updated by sliding forward one step. The attention model is subsequently fine-tuned on the updated window using the same training configuration, enabling real-time adaptation to distributional shifts.

\begin{figure*}[!t]
  \centering

  \begin{subfigure}[!t]{.9\textwidth}
    \centering
    \includegraphics[width=0.42\linewidth]{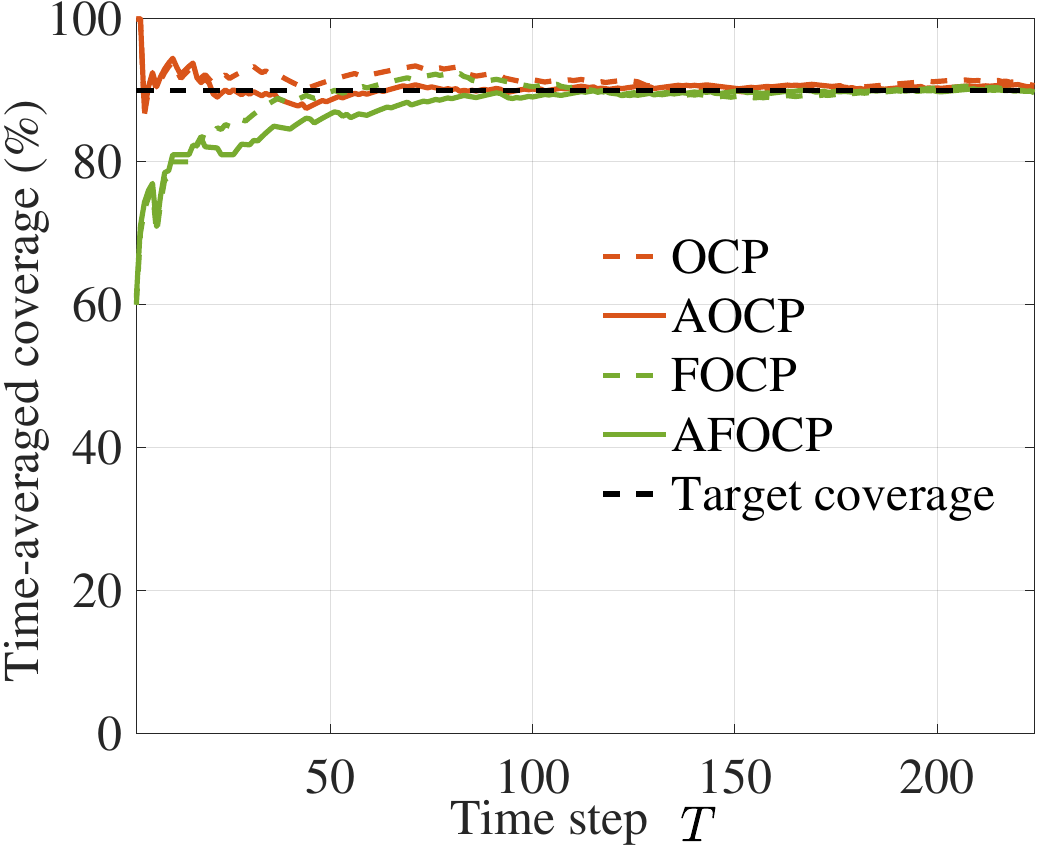}\hspace{.02\linewidth}
    \includegraphics[width=0.42\linewidth]{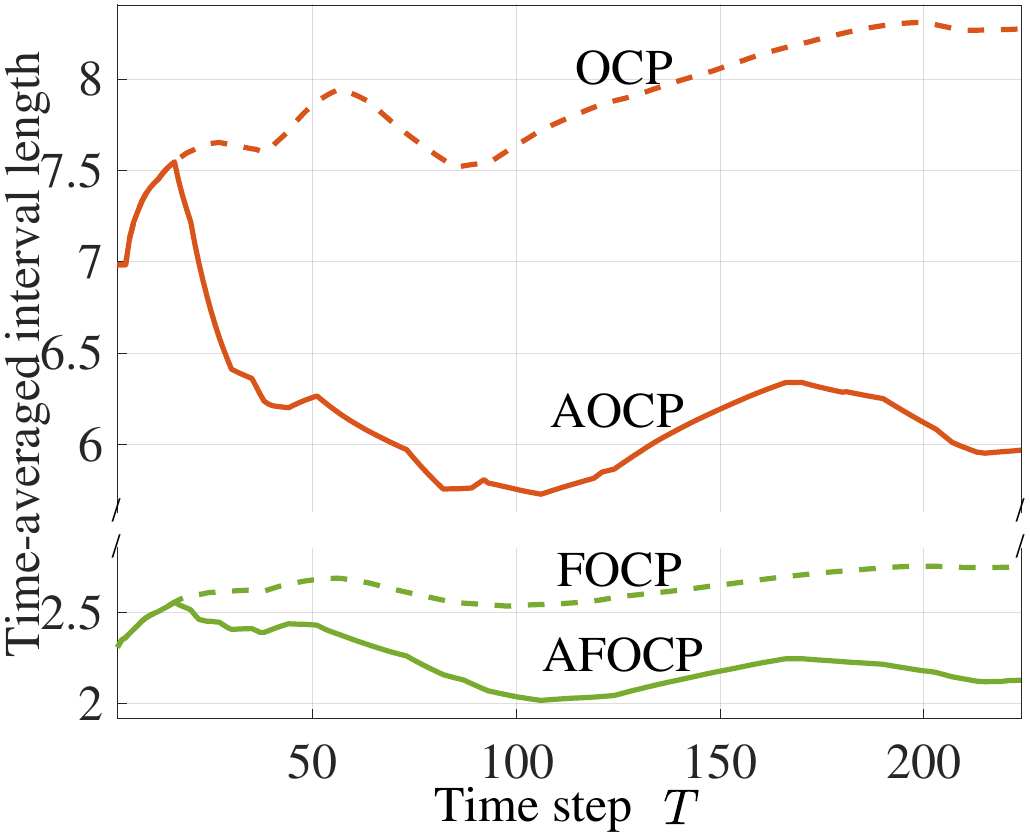}
    \vspace{-4mm}
    \caption[Synthetic data]{\footnotesize Synthetic data}
  \end{subfigure}\par\vspace{-1mm}

  \begin{subfigure}[!t]{.9\textwidth}
    \centering
    \includegraphics[width = 0.43\linewidth]{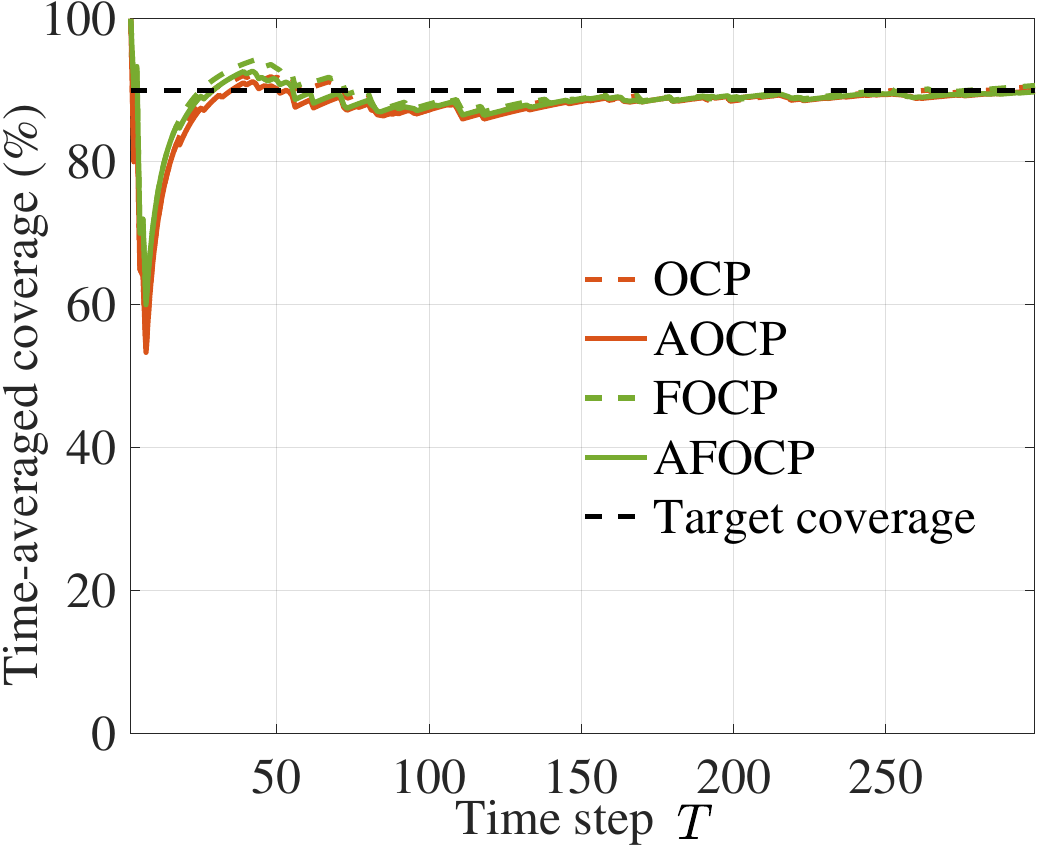}\hspace{.02\linewidth}
    \includegraphics[width=0.42\linewidth]{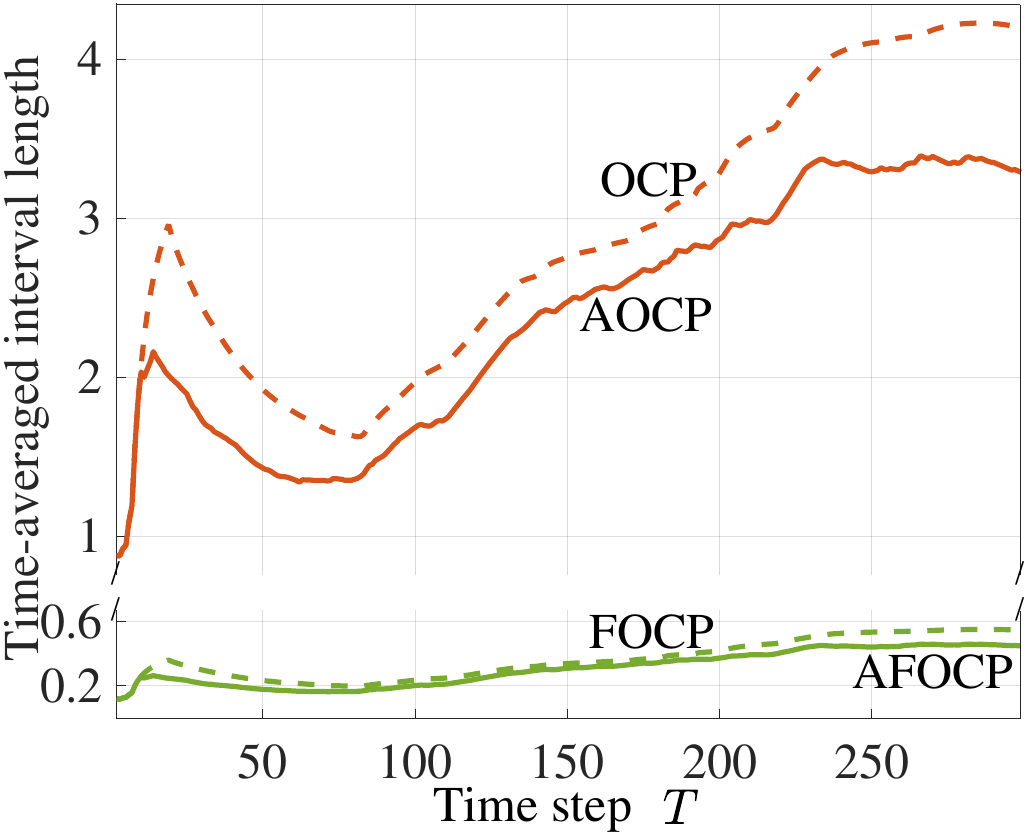}
    \vspace{-4mm}
    \caption[Air quality]{\footnotesize Air quality}
    \end{subfigure}\par\vspace{-1mm}

  \begin{subfigure}[!t]{.9\textwidth}
    \centering
    \includegraphics[width=0.42\linewidth]{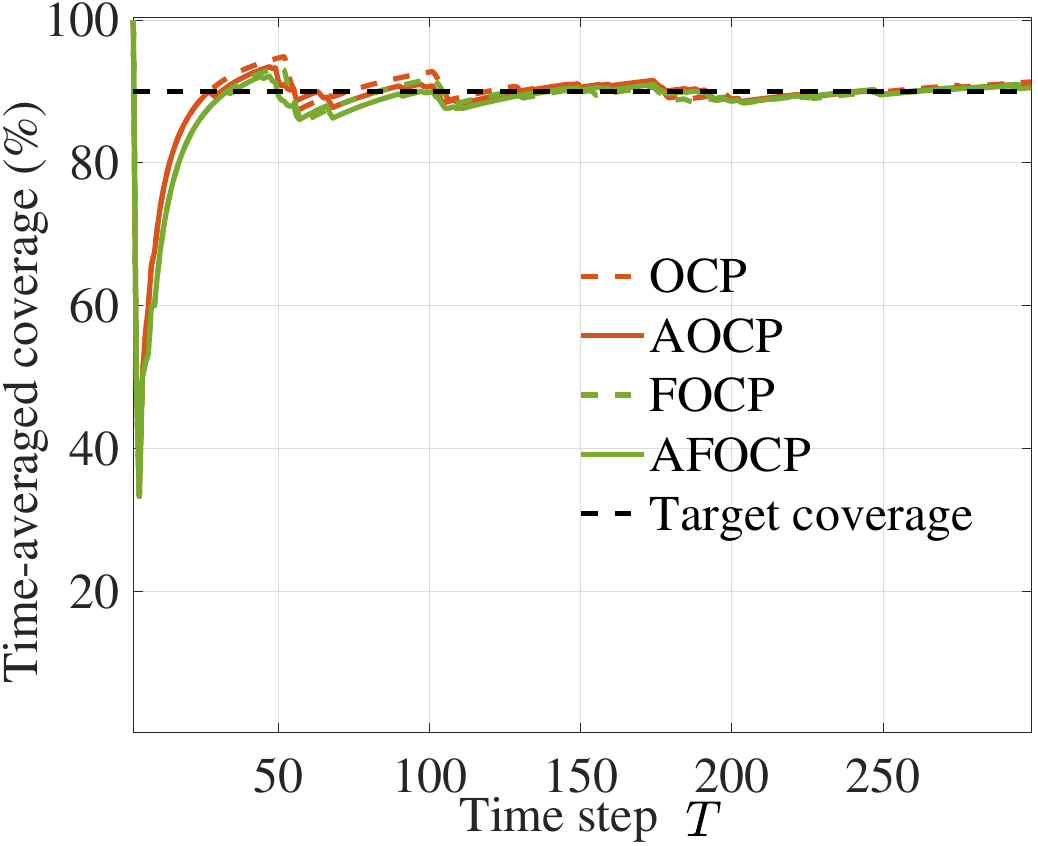}\hspace{.02\linewidth}
    \includegraphics[width=0.42\linewidth]{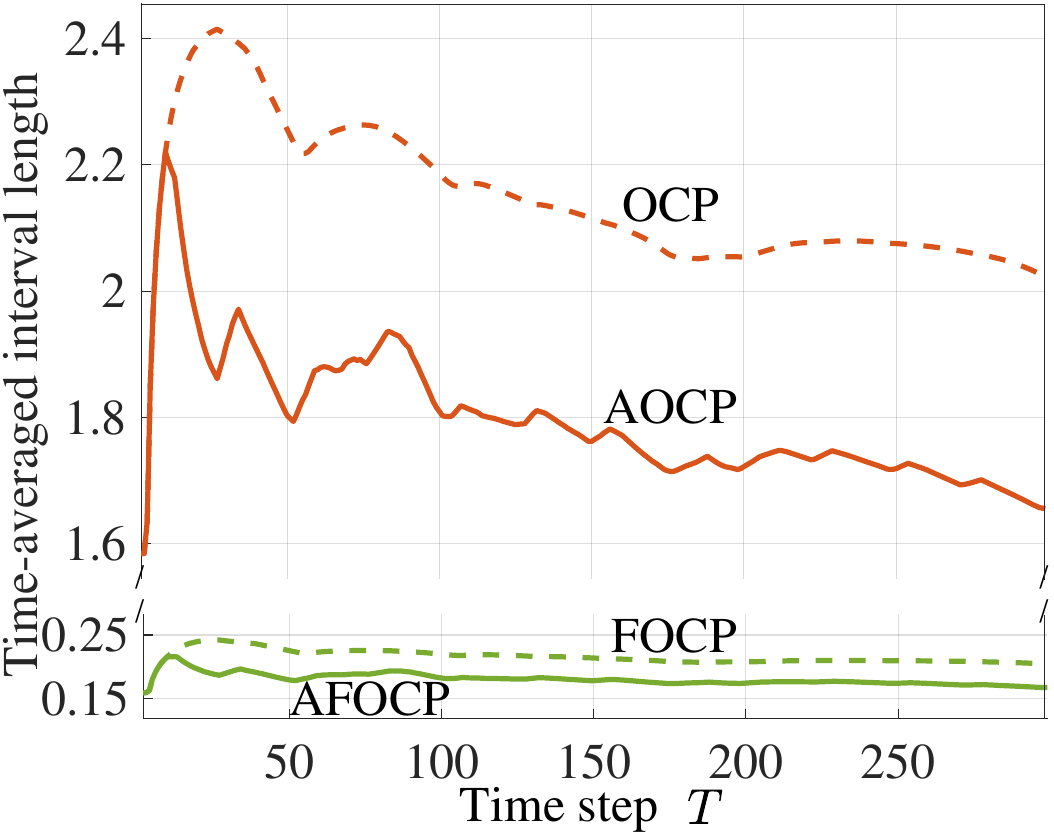}
    \vspace{-4mm}
    \caption[Electricity]{\footnotesize Electricity}
  \end{subfigure}\par\vspace{-1mm}

  \begin{subfigure}[!t]{.9\textwidth}
    \centering
    \includegraphics[width=0.42\linewidth]{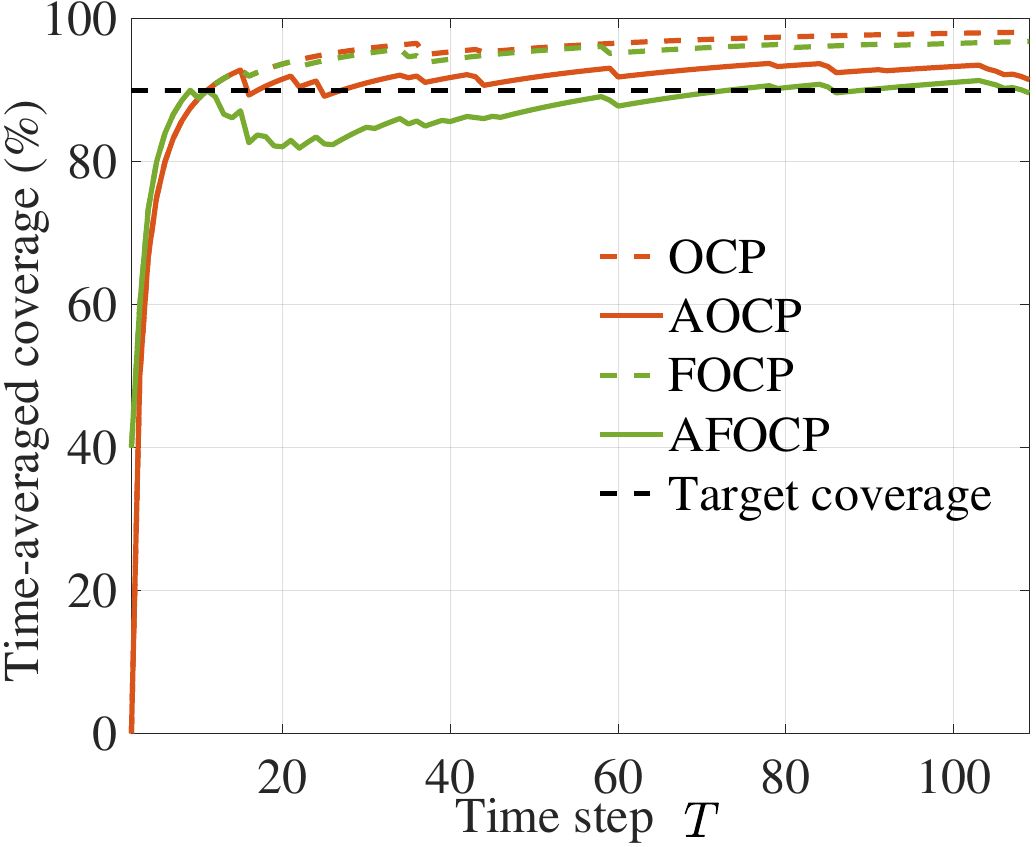}\hspace{.02\linewidth}
    \includegraphics[width=0.42\linewidth]{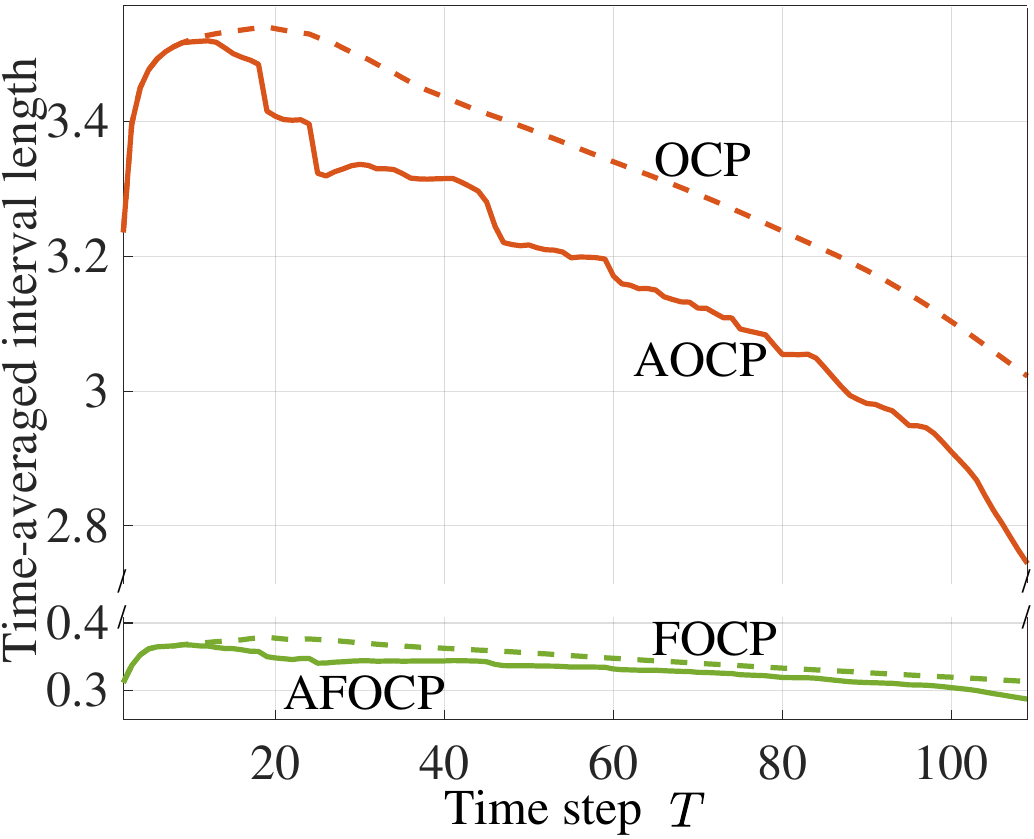}
    \vspace{-4mm}
    \caption[Bike-sharing]{\footnotesize Bike-sharing}
  \end{subfigure}\par\vspace{-1mm}
\end{figure*}

\begin{figure*}\ContinuedFloat
  \centering
  \begin{subfigure}[!t]{.9\textwidth}
    \centering
    \includegraphics[width=0.42\linewidth]{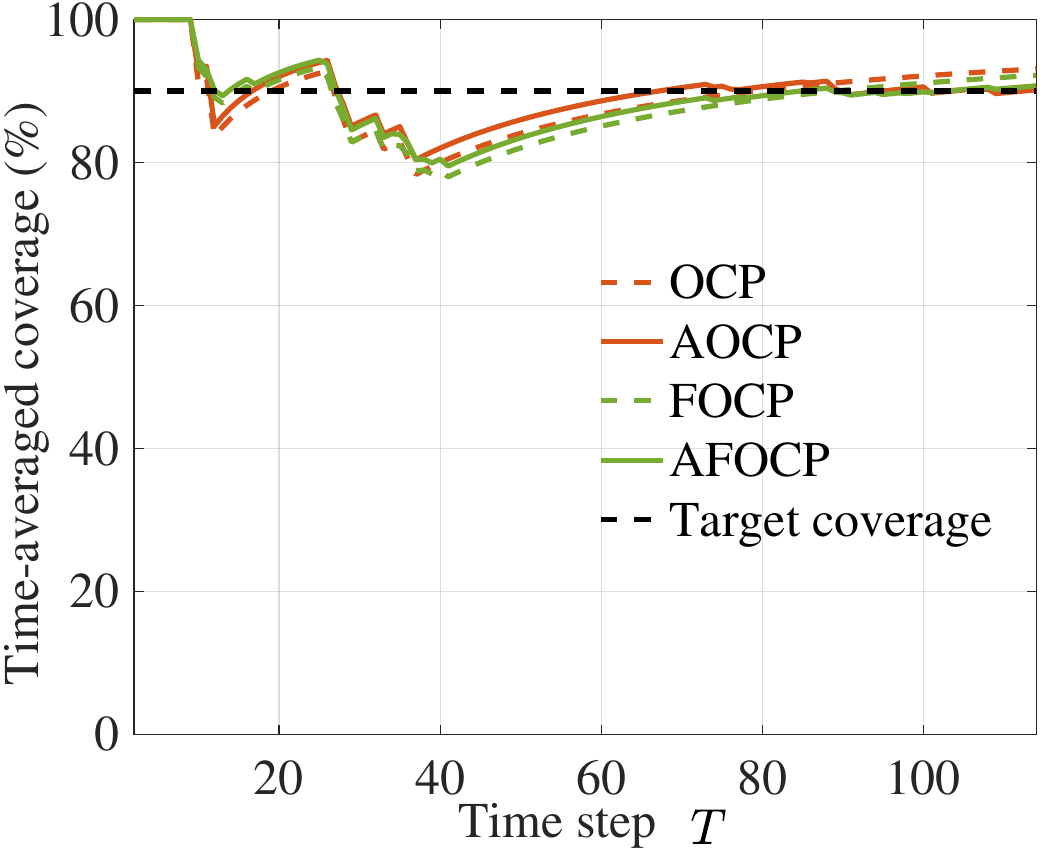}\hspace{.02\linewidth}
    \includegraphics[width=0.42\linewidth]{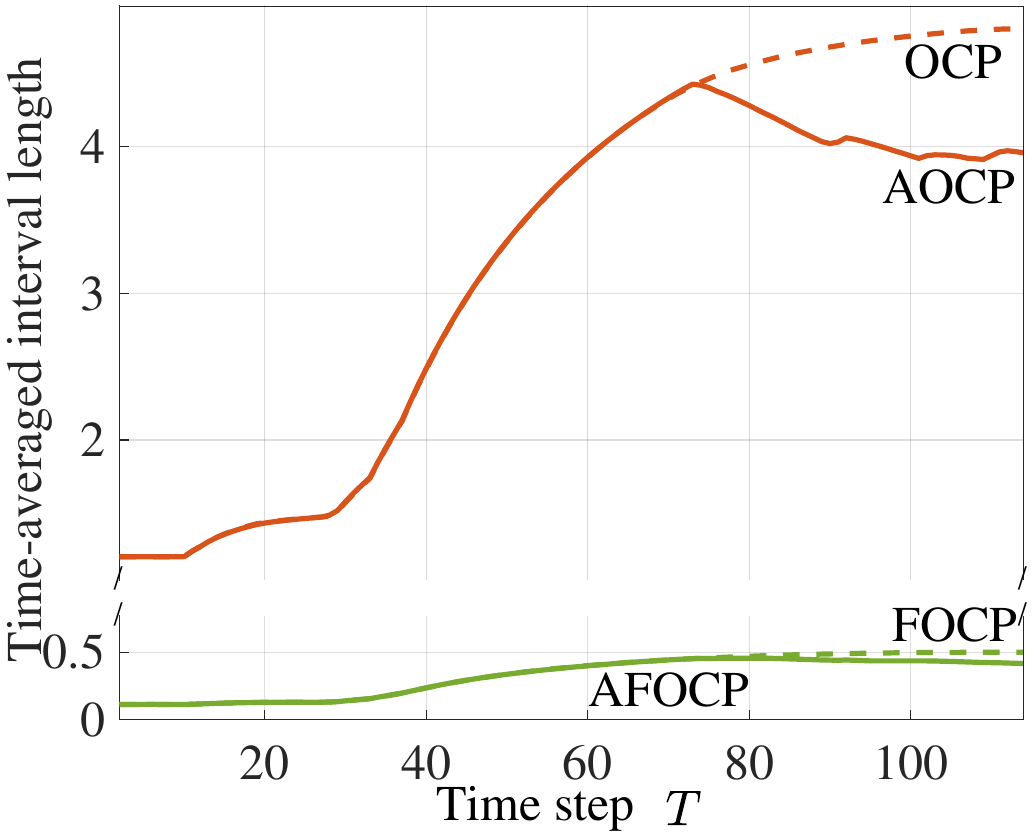}
    \vspace{-4mm}
    \caption[Wind]{\footnotesize Wind}
  \end{subfigure}\par\vspace{-1mm}
 \vspace{-1mm}
  \caption{Time-averaged coverage (left) and time-averaged interval length (right) of OCP, FOCP, AOCP, and AFOCP versus time $T$ across various datasets, with window length $L = 100$, feature dimension $D = 50$, and target miscoverage rate $\alpha=0.1$.}
  \label{time_cov_len}
\end{figure*}

\begin{figure*}[!t]
  \centering

  \begin{subfigure}[!t]{.9\textwidth}
    \centering
    \includegraphics[width=0.42\linewidth]{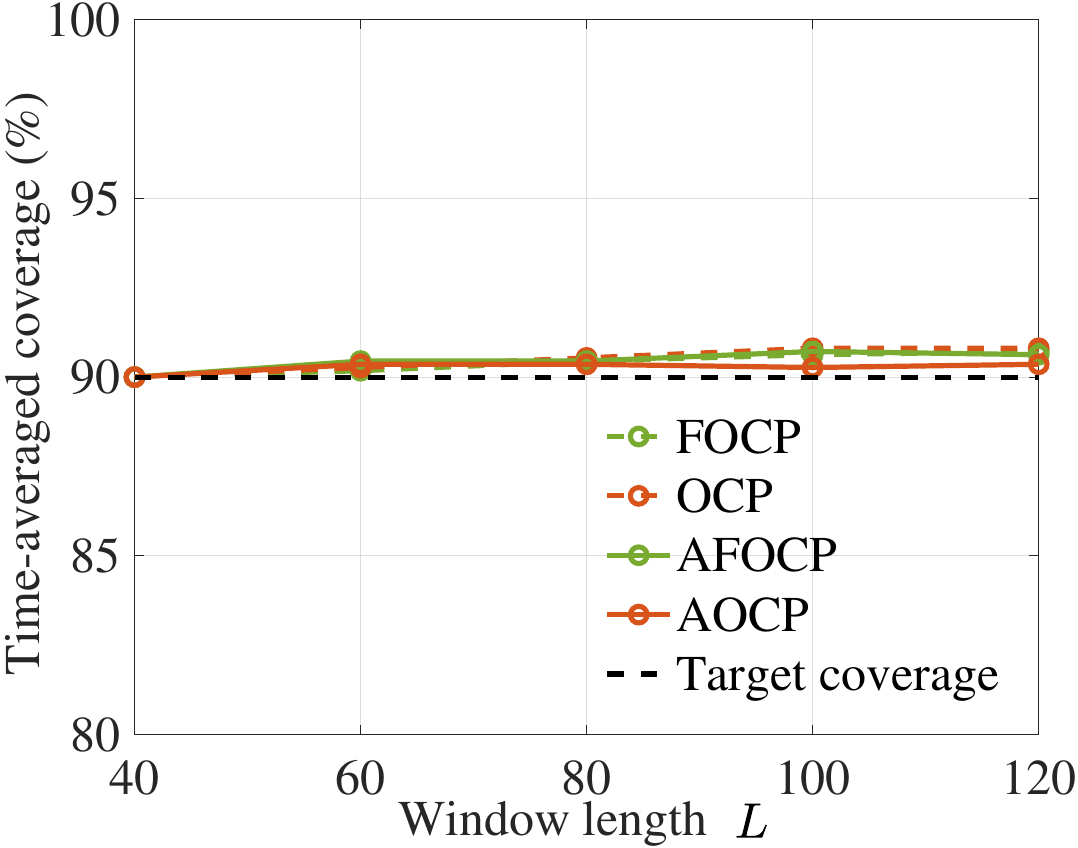}\hspace{.02\linewidth}
    \includegraphics[width=0.42\linewidth]{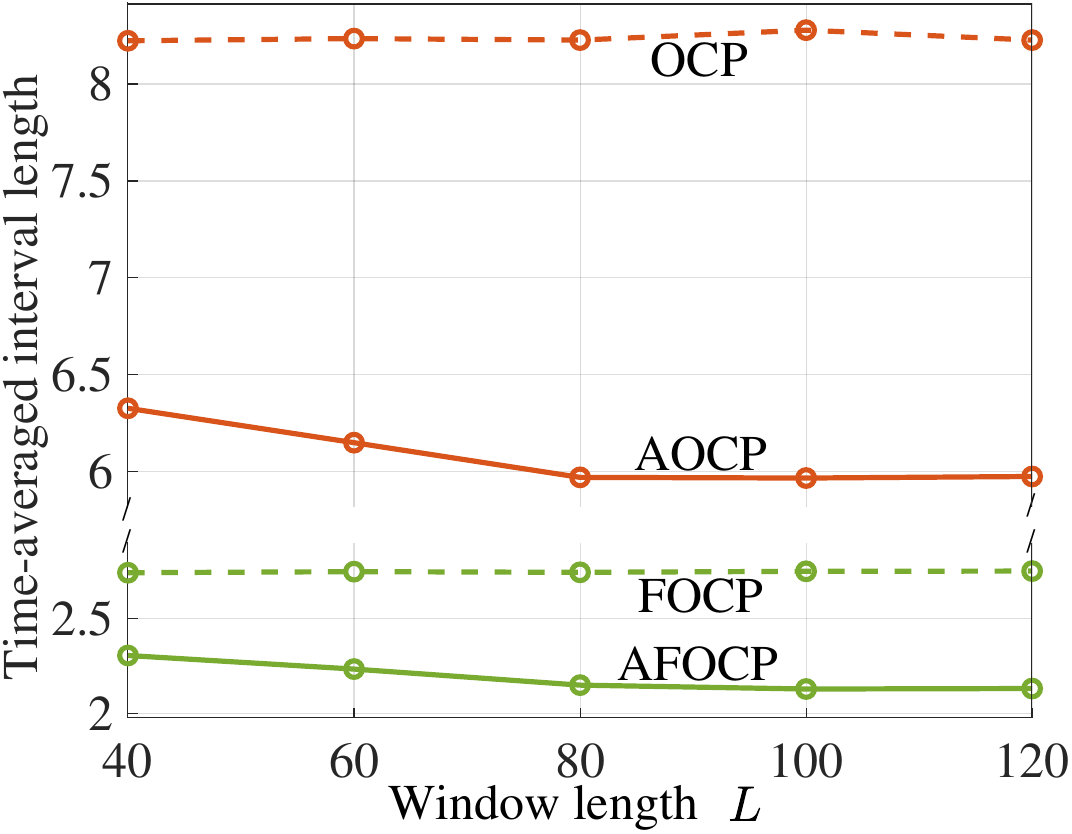}
    \vspace{-4mm}
    \caption[Synthetic data]{\footnotesize Synthetic data}
  \end{subfigure}\par\vspace{-1mm}

  \begin{subfigure}[!t]{.9\textwidth}
    \centering
    \includegraphics[width=0.42\linewidth]{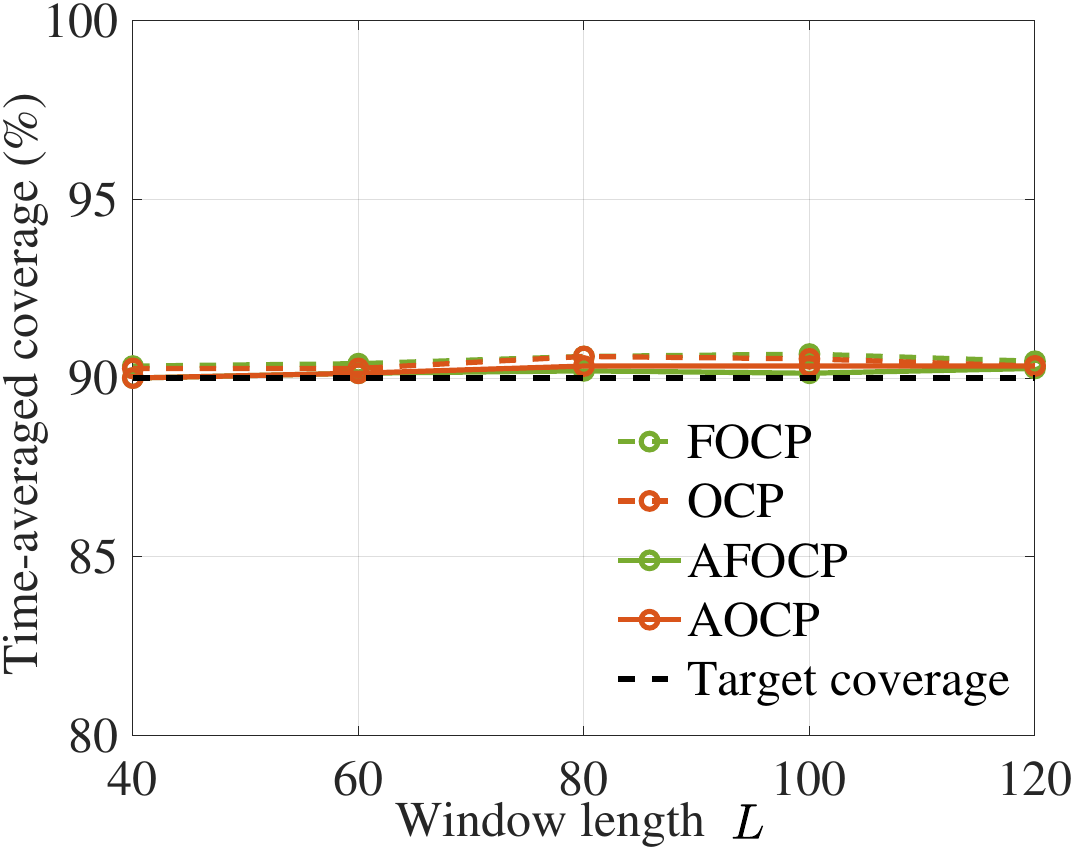}\hspace{.02\linewidth}
    \includegraphics[width=0.42\linewidth]{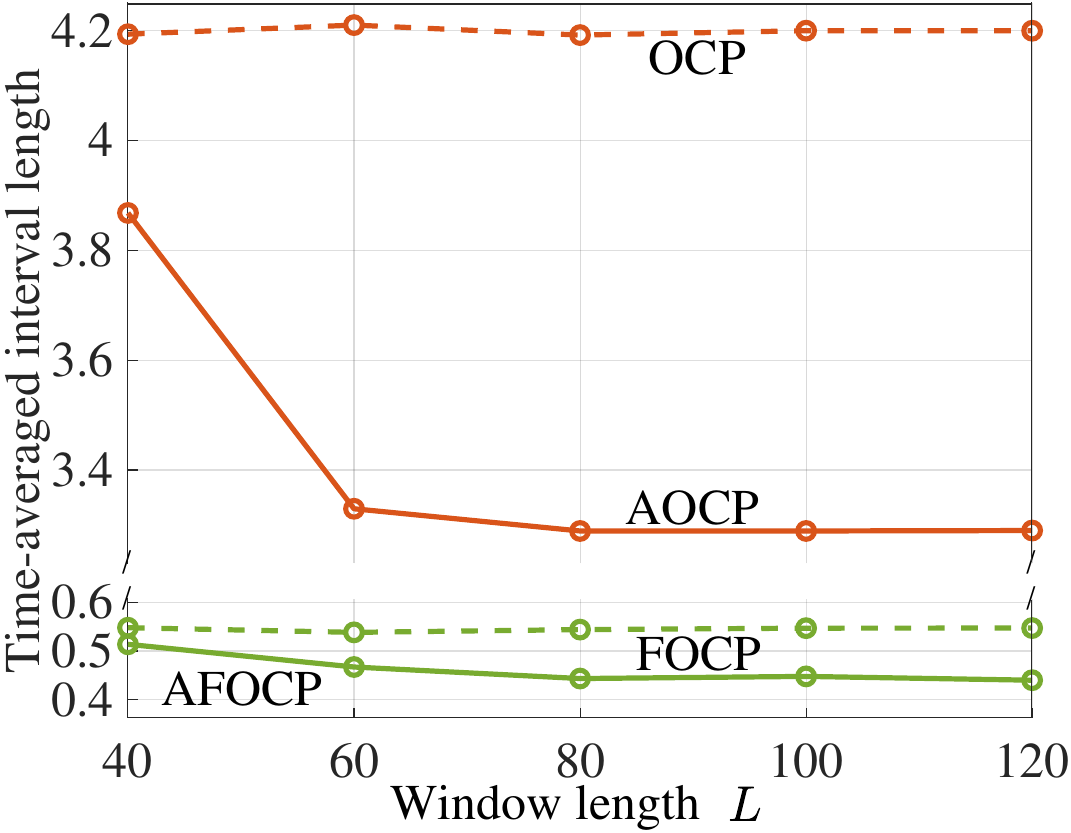}
    \vspace{-4mm}
    \caption[Air quality]{\footnotesize Air quality}
    \end{subfigure}\par\vspace{-1mm}
  \caption{Time-averaged coverage (left) and time-averaged interval length (right) of OCP, FOCP, AOCP, and AFOCP versus window length $L$ for synthetic data and air quality datasets with feature dimension $D= 50$ and target miscoverage rate $\alpha=0.1$.}
  \label{win_cov_len}
\end{figure*}

\begin{figure*}[!t]
  \centering

  \begin{subfigure}[!t]{.9\textwidth}
    \centering
    \includegraphics[width=0.42\linewidth]{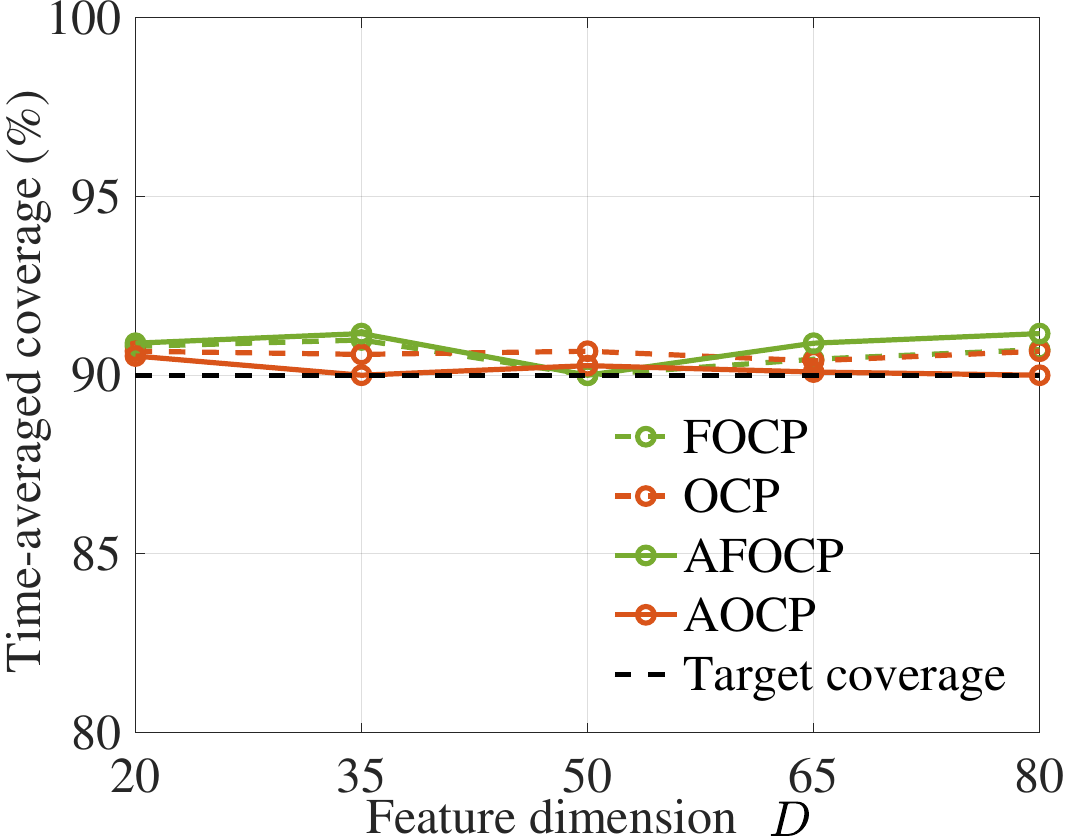}\hspace{.02\linewidth}
    \includegraphics[width=0.41\linewidth]{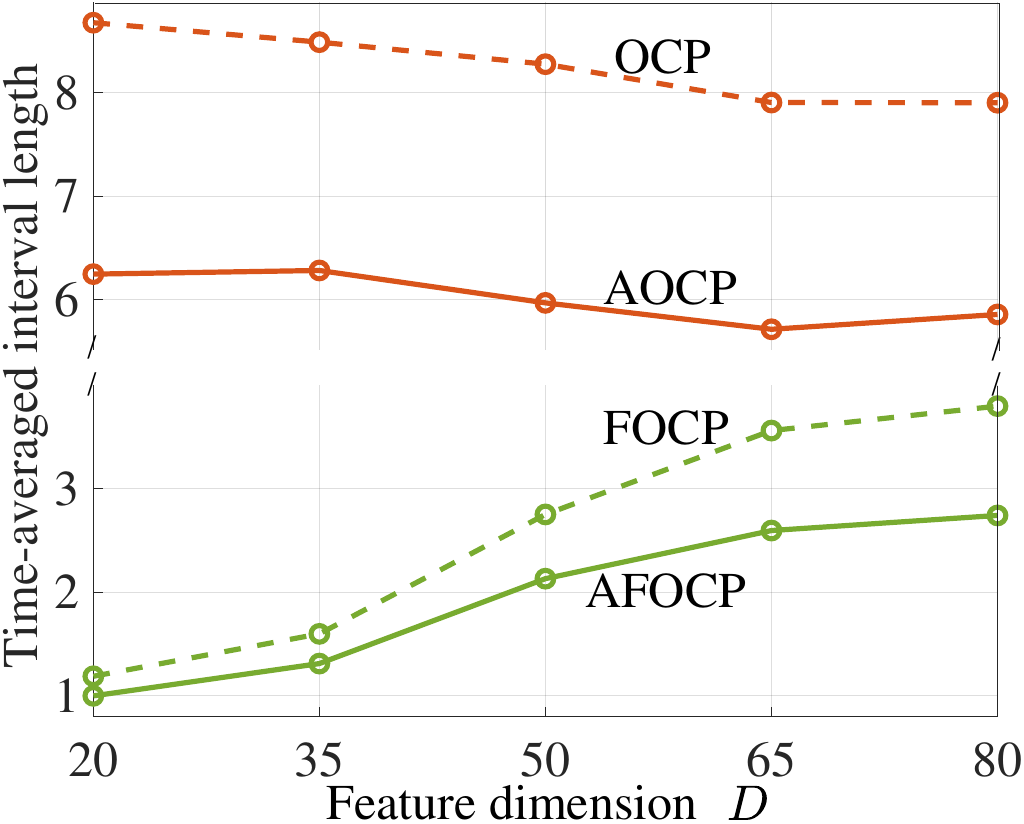}
    \vspace{-4mm}
    \caption[Synthetic data]{\footnotesize Synthetic data}
  \end{subfigure}\par\vspace{-1mm}

  \begin{subfigure}[!t]{.9\textwidth}
    \centering
    \includegraphics[width=0.42\linewidth]{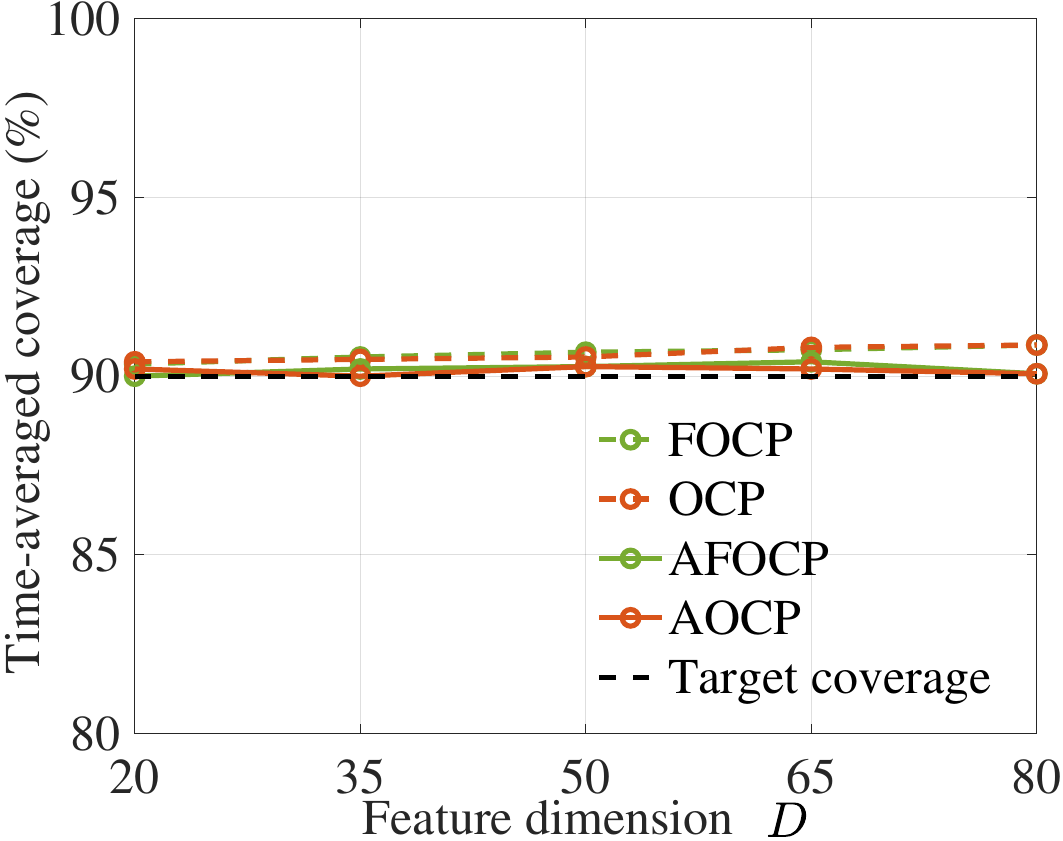}\hspace{.02\linewidth}
    \includegraphics[width=0.42\linewidth]{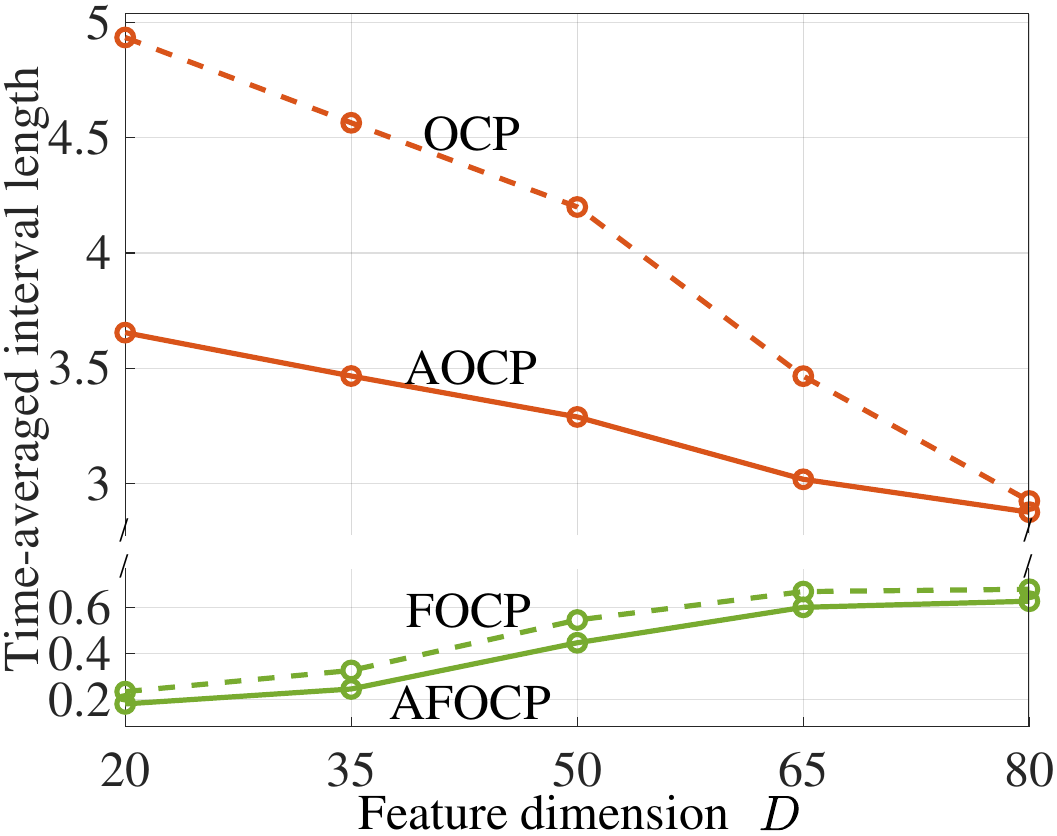}
    \vspace{-4mm}
    \caption[Air quality]{\footnotesize Air quality}
    \end{subfigure}\par\vspace{-1mm}

  \begin{subfigure}[!t]{.9\textwidth}
    \centering
    \includegraphics[width=0.42\linewidth]{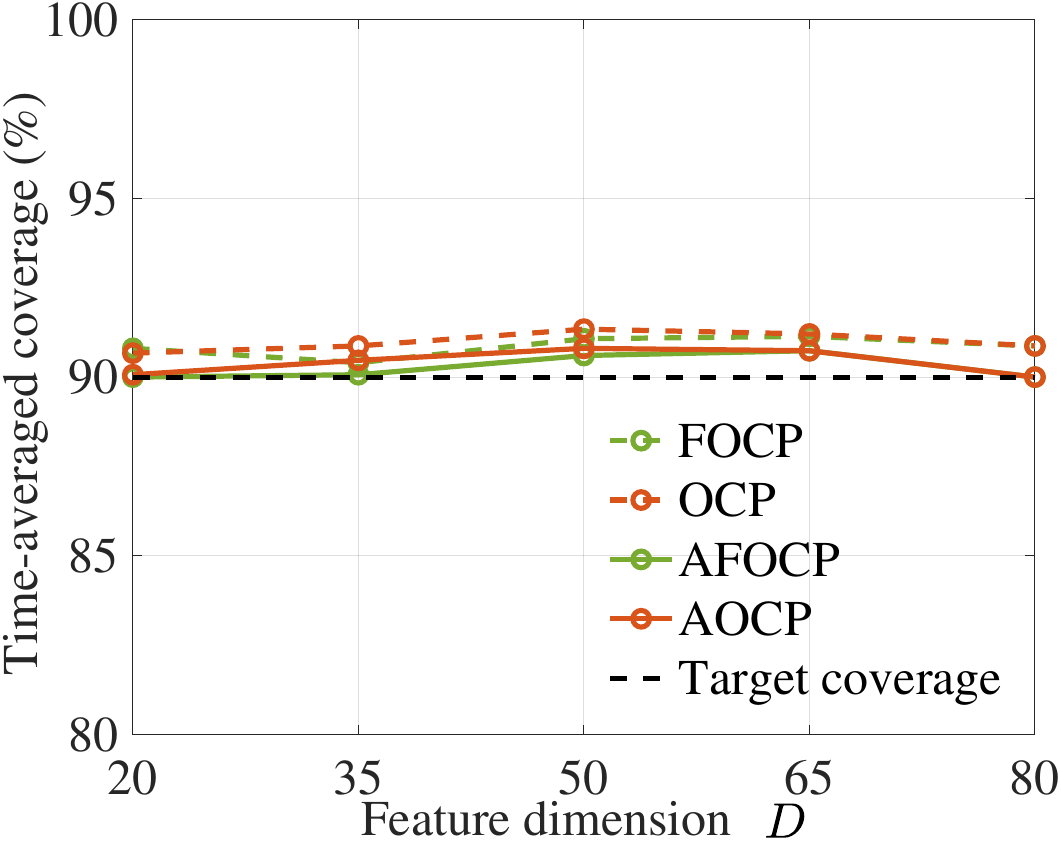}\hspace{.02\linewidth}
    \includegraphics[width=0.42\linewidth]{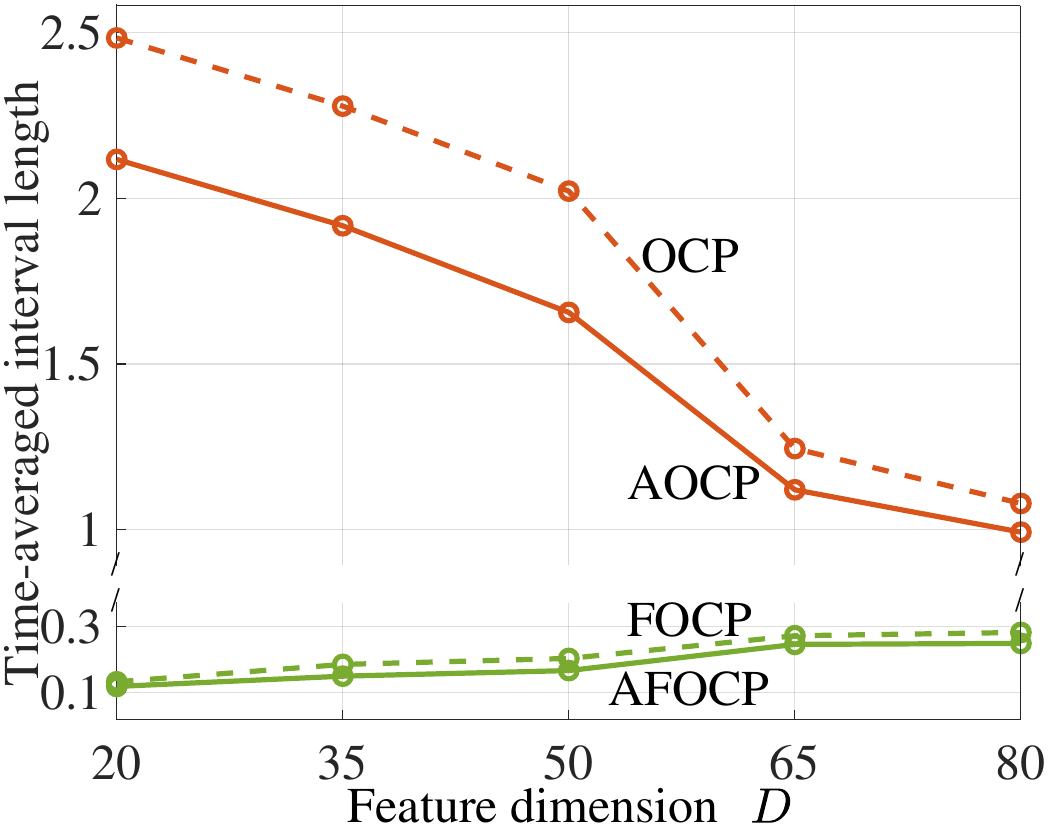}
    \vspace{-4mm}
    \caption[Electricity]{\footnotesize Electricity}
  \end{subfigure}\par\vspace{-1mm}

  \begin{subfigure}[!t]{.9\textwidth}
    \centering
    \includegraphics[width=0.42\linewidth]{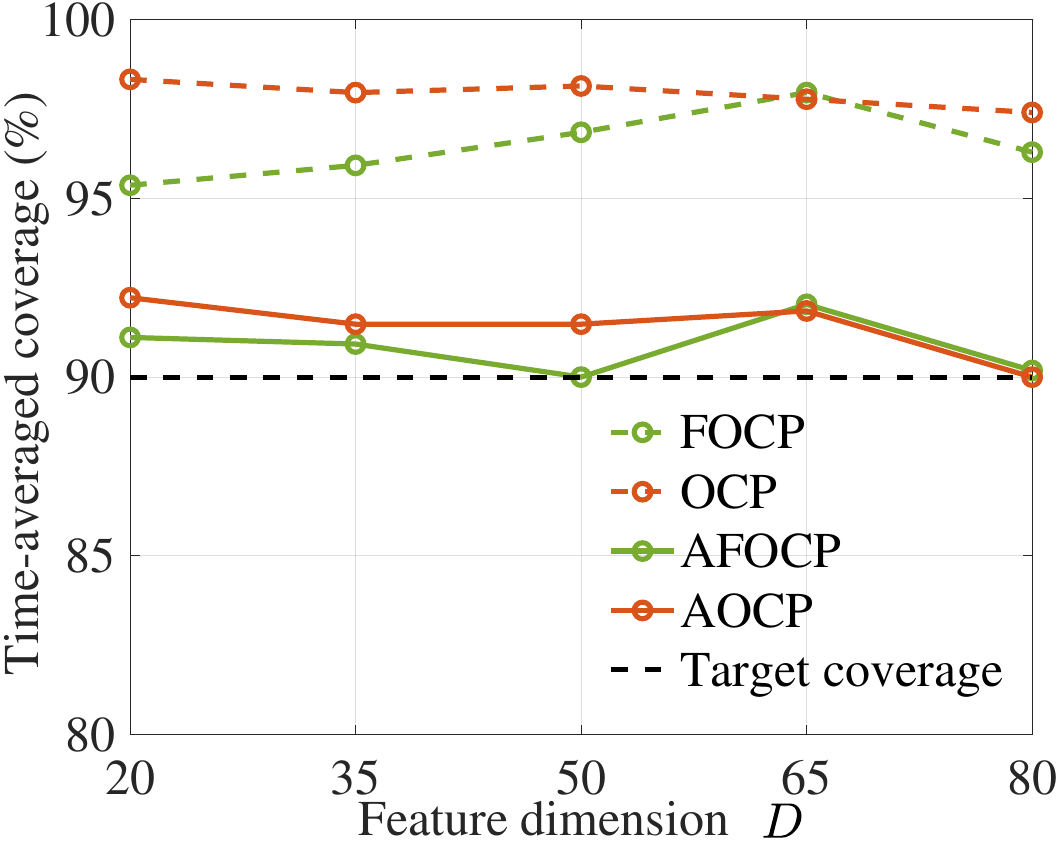}\hspace{.02\linewidth}
    \includegraphics[width=0.42\linewidth]{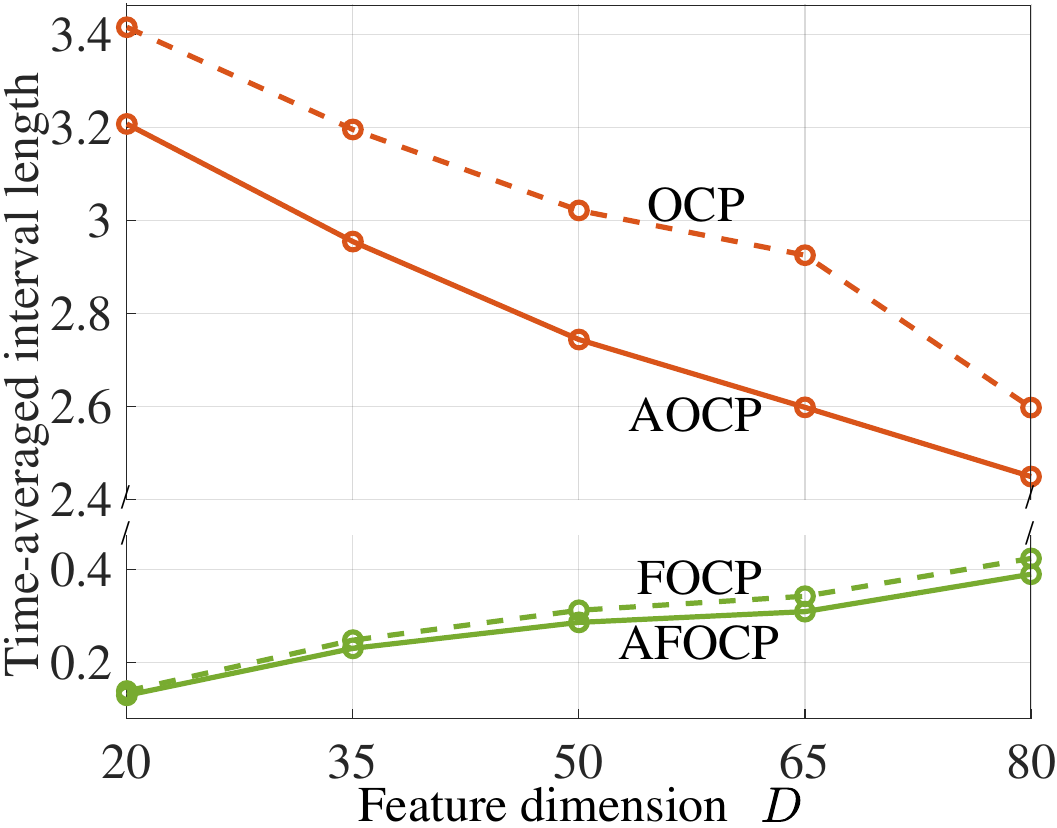}
    \vspace{-4mm}
    \caption[Bike-sharing]{\footnotesize Bike-sharing}
  \end{subfigure}\par\vspace{-1mm}
\end{figure*}

\begin{figure*}\ContinuedFloat
  \centering
  \begin{subfigure}[!t]{.9\textwidth}
    \centering
    \includegraphics[width=0.42\linewidth]{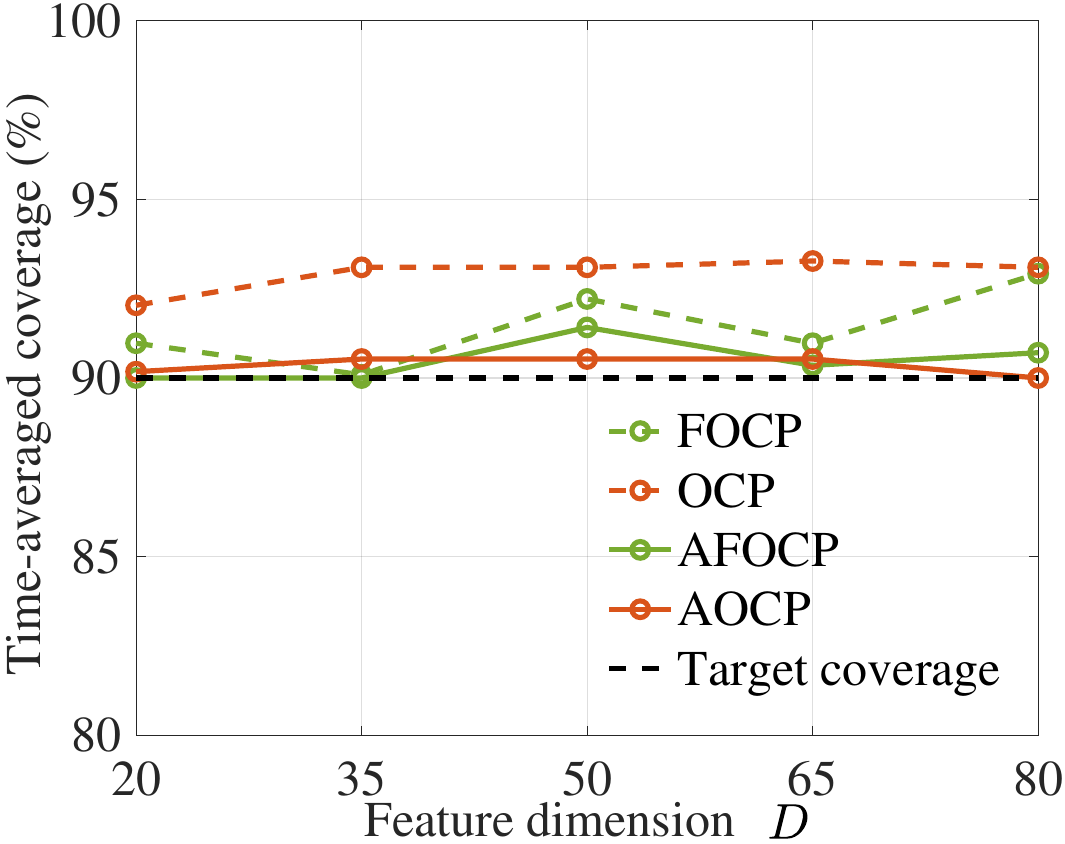}\hspace{.02\linewidth}
    \includegraphics[width=0.42\linewidth]{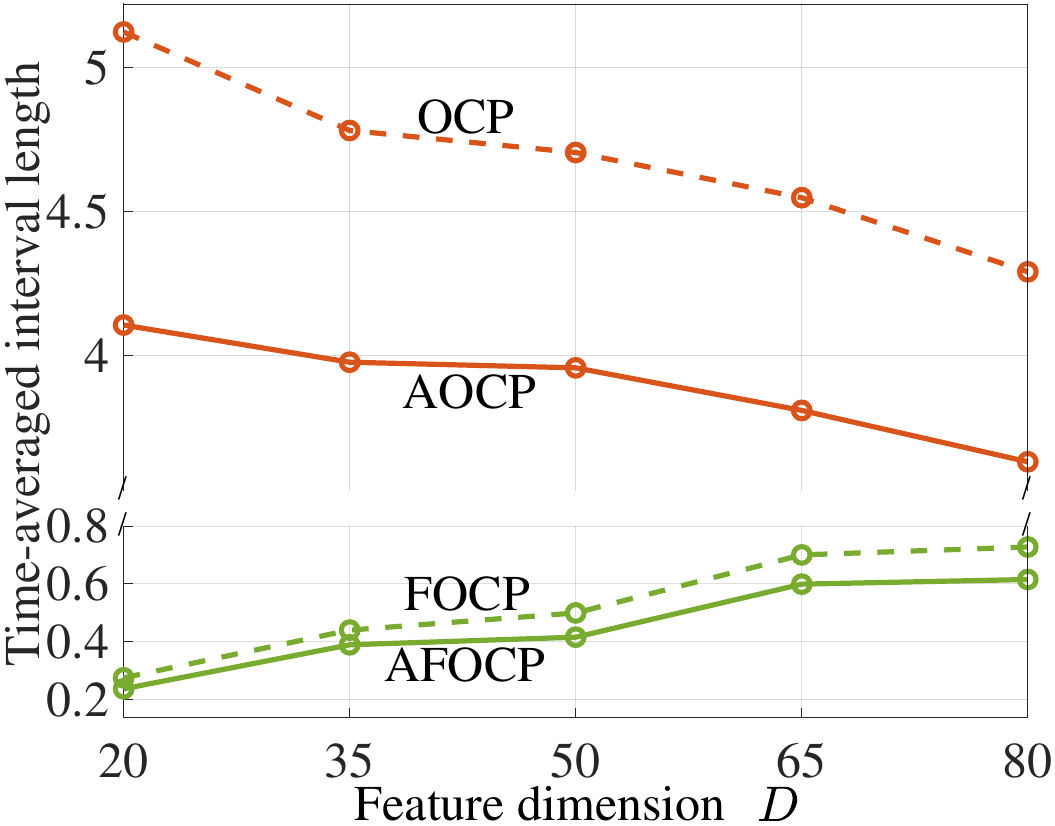}
    \vspace{-4mm}
    \caption[Wind]{\footnotesize Wind}
  \end{subfigure}\par\vspace{-1mm}

  \caption{Time-averaged coverage (left) and time-averaged interval length (right) of OCP, FOCP, AOCP, and AFOCP versus feature dimension $D$ across various datasets with window length $L=100$ and target miscoverage rate $\alpha=0.1$.}
  \label{feat_dim_cov_len}
\end{figure*}

\subsubsection{Evaluation metrics}
We adopt two standard metrics in OCP: time-averaged coverage and time-averaged prediction interval length, which assess reliability and efficiency, respectively.

Consider a multi-dimensional response $Y_t = (Y_t^{(1)}, \ldots,Y_t^{(i)}, \ldots, Y_t^{(I)})\in\mathbb{R}^I$ and its prediction set $\Gamma^*_t \left(X_t\right) \subseteq\mathbb{R}^I$, where the length along each dimension forms a vector $\left|\Gamma^*_t\left(X_t\right)\right|\in\mathbb{R}^I$, with $*\in\left\{\textrm{OCP},\textrm{AOCP},\textrm{FOCP},\textrm{AFOCP}\right\}$. The two metrics are defined as
\begin{subequations}
    \begin{align}
        \textrm{Time-averaged coverage} &= \frac{1}{T}\sum_{t=1}^T\mathds{1}\left\{Y_t\in\Gamma^*_t\left(X_t\right)\right\},\label{subeq_cov}\\
        \textrm{Time-averaged interval length} &= \frac{1}{T}\sum_{t=1}^T\frac{1}{I}\sum_{i=1}^I\left|\Gamma^*_t\left(X_t\right)\right|_{(i)},\label{subeq_len}
    \end{align}
\end{subequations}
where the subscript $(i)$ in \eqref{subeq_len} denotes the $i$-th dimension of vector $\Gamma^*_t(X_t)$, i.e., the prediction interval corresponding to response $Y_t^{(i)}$. The indicator function $\mathds{1}\left\{\cdot\right\}$ in \eqref{subeq_cov} takes $1$ if all $I$ entries of $Y_t$ lie within the prediction interval, and \eqref{subeq_len} also averages interval length across dimensions.

\subsection{Performance Evaluation}
We evaluate OCP, AOCP, FOCP, and AFOCP on the five datasets described in Sec. \ref{sec_data_set} by showing the time-averaged coverage in \eqref{subeq_cov} and the time-averaged interval length in \eqref{subeq_len} as a function of time $T$ in Fig. \ref{time_cov_len}. For all datasets, we choose window length as $L = 100$, the feature dimension as $D = 50$, and the target miscoverage rate as $\alpha = 0.1$ (dashed line). Across datasets, the long-term coverage of all methods converges to the target level, confirming the theoretical results in Corollary \ref{coro_on_fcp}. In terms of efficiency, feature-space calibration (FOCP and AFOCP) yields substantially shorter time-averaged interval lengths than output-space calibration (OCP and AOCP), highlighting the advantage of operating in the space of task-aligned representations. Incorporating attention for adaptive weighting (AOCP and AFOCP) further reduces interval lengths compared to uniform weighting (OCP and FOCP). The latter gain is particularly pronounced for the synthetic dataset, which presents a higher degree of non-stationarity as compared to the other datasets.


Fig. \ref{win_cov_len} and Fig. \ref{feat_dim_cov_len} report the performance at time $T$ equal to the test sequence length for each dataset of OCP, AOCP, FOCP, and AFOCP versus the window length $L$ with $D = 50$ and $\alpha = 0.1$, and versus the feature dimension $D$ with $L = 100$ and $\alpha = 0.1$. For the window-length analysis, we include only the synthetic and air-quality datasets, which share a similar structure in alternating segments, making the results easier to interpret (see Sec. \ref{sec_data_set}). As seen in the figures, coverage remains close to the nominal level across all settings. Furthermore, feature-space calibration (FOCP and AFOCP) consistently achieves shorter converged intervals than output-space calibration (OCP and AOCP), while attention-based weighting (AOCP and AFOCP) provides additional reductions over each uniform weighting baseline (OCP and FCOP).

As also shown in Fig. \ref{win_cov_len}, increasing the window size $L$ decreases the time-averaged interval length for AOCP and AFOCP up to $L = 80$, after which further gains are negligible. This threshold coincides with the maximum segment length over which the process is approximately stationary, as defined in Sec. \ref{sec_data_set}. Once the window covers a full segment, adding more history provides minimal gains. The use of attention is particularly important when the window tends to contain a mix of different segments, i.e., when $L>80$, as in this case, attention can learn to assign larger weights to features within the same segment.

Finally, as shown in Fig. \ref{feat_dim_cov_len}, increasing the feature space dimension $D$ widens the intervals for feature-space calibration (FOCP and AFOCP), but narrows them for output-space calibration (OCP and AOCP). In the former case, a larger dimension $D$ has the dominant effect of introducing additional nuisance variations and amplifying the approximation error in the inverse mapping used for feature-level NC scores, thus increasing score dispersion and the resulting quantiles. In the latter case, a larger dimension $D$ benefits performance due to improved predictive fit of the underlying model, reducing residual errors and thus the output-level quantiles.

\section{Conclusions}\label{sec_conclusion}
This work introduced AFOCP, a principled extension of OCP that calibrates uncertainty in the learned feature space and adaptively reweights historical observations via an attention mechanism. By shifting calibration from the output space to task-aligned representations and replacing uniform aggregation with relevance weights, AFOCP reduces nuisance variation to concentrate NC scores in feature space and adapts the quantile to the current test point. The result is a simple, modular, and effective approach to reliable and efficient uncertainty quantification for non-stationary time series.

We established two guarantees. First, AFOCP attains the target long-term coverage. Second, under mild regularity conditions, AFOCP yields shorter time-averaged prediction intervals than its output-space counterpart.
Experiments on synthetic and real-world time series support the theory: AFOCP maintains nominal coverage while substantially reducing prediction interval length relative to OCP. The intermediate variants FOCP and AOCP further isolate the contributions of feature-space calibration and attention-based weighting.

Future work includes richer attention architectures such as multi-head and multi-scale designs, adaptive selection of calibration history, and more efficient streaming implementations. Furthermore, extending the analysis to broader model classes, weaker assumptions, structured outputs, and multivariate coverage criteria is a promising direction.

\appendix

\subsection{Formal Description and Proof of Theorem \ref{theo_on_fcp_band}}\label{sec_proof}
Our conclusions build on Theorem 6 in \cite{teng2022predictive} and Theorem 1 in \cite{chenconformalized}. We start by presenting key definitions for AOCP and AFOCP.

At time $t$, we use the most recent $L$ observed data pairs $\left\{(X_{\tau}, Y_{\tau})\right\}_{\tau = t-L}^{t-1}$ to compute NC scores, together with non-negative normalized weights $\left\{w^{\tau}_t\right\}_{\tau=1}^{L+1}$ for weighted quantile calculation.

\textbf{AOCP.}
Let $M_{t-L:t-1}=\left\{M_\tau\right\}_{\tau = t-L}^{t-1}$ denote the individual lengths in the output space for the latest $L$ data pairs, where $M_{\tau} := 2s(X_\tau,Y_\tau)$ and $s(X_{\tau},Y_{\tau})$ is the NC score defined in \eqref{eq_NC}. Thus, the prediction interval length is $Q_{1-\alpha_t} (M_{t-L: t-1})$, where $Q_{1-\alpha_t} (M_{t-L: t-1})$ is the $(1-\alpha_t)$-quantile of the weighted empirical distribution $\sum_{\tau=1}^{L} w_t^{\tau} \delta_{M_{t-\tau}} + w_t^{L+1} \delta_{+\infty}$, where $\delta_b$ is the point mass as in \eqref{eq_CP_pred_set}.

\textbf{AFOCP.} Let $M^\textrm{f}_{t-L:t-1} = \left\{M^\textrm{f}_\tau\right\}_{\tau = t-L}^{t-1}$ denote the individual lengths in the feature space for the latest $L$ data pairs, where $M^\textrm{f}_\tau:=2s^\textrm{f}(X_\tau,Y_\tau)$ and $s^\textrm{f}(X_\tau,Y_\tau)$ is the feature-based NC score defined in \eqref{eq_fcp_NC}. To characterize the prediction length in the output space, we define $\mathcal{H}\left(M, X\right)$ as the individual output-space length associated with input $X$ given feature-space length $M$. Specifically, $\mathcal{H}\left(M,X\right)$ represents the length of the set $\left\{g(U):\left\|U-f(X)\right\|\leq M/2\right\}$, which maps the diameter $M$, centered at $f(X)$, through the prediction head $g(\cdot)$ and returns the maximal spread of the resulting outputs. Accordingly, the prediction interval length of AFOCP is $\mathcal{H}\big(Q_{1-\alpha_t} (M^\textrm{f}_{t-L: t-1}), X_t\big)$, where $Q_{1-\alpha_t} (M^\textrm{f}_{t-L: t-1})$ calculates the $(1-\alpha_t)$-quantile of the weighted empirical distribution $\sum_{\tau=1}^{L} w_{t}^{\tau} \delta_{M^{\textrm{f}}_{t-\tau}} + w_{t}^{L+1} \delta_{+\infty}$, where $\delta_b$ is the point mass as in \eqref{eq_CP_pred_set}. Without abuse of notations, operating $\mathcal{H}$ on a dataset means operating $\mathcal{H}$ on each data point in the set, e.g., $\mathcal{H}(M^\textrm{f}_{t-L: t-1}, X_{t-L: t-1})=\big\{\mathcal{H}(M^\textrm{f}_\tau, X_\tau)\big\}_{\tau=t-L}^{t-1}$.

Following the structure of \cite{teng2022predictive, chenconformalized}, we then present the formal and mathematical description of Theorem \ref{theo_on_fcp_band}.

\begin{theorem}\label{theo_on_fcp_band2}
    Assume the inequality that there exist constants $\beta>0$ and $R>0$ satisfying
    \begin{align}\label{eq_hold_org}
        \left|\mathcal{H}(M,X)-\mathcal{H}(M',X) \right|\leq R\left|M-M'\right|^{\beta}
    \end{align}
    for all $X$. Additionally, assume that there exist constants $\epsilon>0$, $C>0$ such that for all horizons $T\geq 1$, the following time-averaged assumptions hold:
    \begin{itemize}
        \item[1)]\textbf{Length Preservation:} The output-space quantiles induced by the feature-space construction are not significantly larger than the corresponding quantiles based directly on the output-space lengths, namely,
        \begin{align}\label{eq_len_pre}
             \frac{1}{T}\sum_{t=1}^T Q_{1-\alpha_t}\big(\mathcal{H}(M^\text{\rm{f}}_{t-L: t-1}, X_{t-L:t-1})\big) < \frac{1}{T}\sum_{t=1}^T Q_{1-\alpha_t}(M_{t-L: t-1}) + \epsilon
        \end{align}

        \item [2)] \textbf{Expansion:} The operator $\mathcal{H}\left(M,X\right)$ expands the differences between individual lengths and their quantiles, namely,
        \begin{align}\label{eq_expand}
            R & \cdot \frac{1}{T}\sum_{t=1}^T \mathbb{M} \Big[\left|Q_{1-\alpha_t}(M^{\text{\rm{f}}}_{t-L:t-1}) - M^{\text{\rm{f}}}_{t-L:t-1}\right|^{\beta}\Big] \nonumber\\
            & < \frac{1}{T}\sum_{t=1}^T \mathbb{M} \Big[Q_{1-\alpha_t} \big(\mathcal{H}(M^{\text{\rm{f}}}_{t-L:t-1}, X_{t-L: t-1})\big) - \mathcal{H}(M^{\text{\rm{f}}}_{t-L: t-1}, X_{t-L: t-1})\Big] \nonumber \\
            & \hspace{6cm}- \epsilon - 2\max\left\{R,1\right\}\left(\frac{C}{\sqrt{L}}\right)^{ \min\left\{\beta,1\right\}},
        \end{align}
        where $\mathbb{M}[\cdot]$ denotes the mean of a set.

        \item [3)] \textbf{Quantile Stability:} The quantile of the individual length is stable from feature space to output space, namely,
        \begin{align}\label{eq_quant_stab}
            \frac{1}{T} \sum_{t=1}^T   \Big|\mathcal{H} \big(Q_{1-\alpha_t} (M^{\text{\rm{f}}}_{t-L: t-1}), X_{t}\big) - \mathbb{M}\Big[\mathcal{H}\big(Q_{1-\alpha_t}(M^{\text{\rm{f}}}_{t-L: t-1}), X_{t-L: t-1}\big)\Big]\Big| \leq \frac{C}{\sqrt{L}}
        \end{align}
    \end{itemize}
     Then AFOCP provably outperforms AOCP in terms of time-averaged prediction interval length, namely,
    \begin{align}\label{eq_len_con}
        \frac{1}{T} \sum_{t=1}^T \mathcal{H}\big(Q_{1-\alpha_t} (M^{\text{\rm{f}}}_{t-L: t-1}), X_t\big) < \frac{1}{T}\sum_{t=1}^T Q_{1-\alpha_t}\left(M_{t-L: t-1}\right).
    \end{align}
\end{theorem}

\textit{Proof:} By the expansion assumption in \eqref{eq_expand}, we rewrite it as
\begin{align}\label{eq_expand2}
    \frac{1}{T}&\sum_{t=1}^T \mathbb{M}\big[\mathcal{H}(M^{\text{\rm{f}}}_{t-L:t-1}, X_{t-L: t-1})\big] + R \cdot \frac{1}{T}\sum_{t=1}^T \mathbb{M} \left[\left|Q_{1-\alpha_t}(M^{\textrm{f}}_{t-L:t-1}) - M^{\textrm{f}}_{t-L:t-1} \right|^{\beta}\right] \nonumber\\
    &< \frac{1}{T}\sum_{t=1}^T Q_{1-\alpha_t}\big(\mathcal{H}(M^{\textrm{f}}_{t-L: t-1}, X_{t-L: t-1})\big) -\epsilon- 2\max\left\{R,1\right\}\left(\frac{C}{\sqrt{L}}\right)^{\min\left\{\beta,1\right\}}.
\end{align}

By setting $M = Q_{1-\alpha_t}(M^{\textrm{f}}_{t-L:t-1})$, $M' = M^{\textrm{f}}_{t-L:t-1}$ and $X = X_{t-L: t-1}$ in \eqref{eq_hold_org} of the H\"older condition, taking mean $\mathbb{M}[\cdot]$ on both sides and accumulating over $T$ time steps, the following inequality holds
\begin{align}\label{eq_hold}
    \frac{1}{T} & \sum_{t=1}^T \mathbb{M}\Big[\mathcal{H} \big(Q_{1-\alpha_t}(M^{\textrm{f}}_{t-L: t-1}), X_{t-L: t-1}\big) - \mathcal{H}(M^{\textrm{f}}_{t-L: t-1}, X_{t-L: t-1})\Big] \nonumber\\
    &\leq R \cdot \frac{1}{T}\sum_{t=1}^T \mathbb{M} \left[\left|Q_{1-\alpha_t} \left(M^{\textrm{f}}_{t-L: t-1}\right) - M^{\textrm{f}}_{t-L: t-1}\right|^{\beta}\right].
\end{align}
We rewrite it as
\begin{align}\label{eq_hold2}
    \frac{1}{T}&\sum_{t=1}^T\mathbb{M}\Big[ \mathcal{H}\big(Q_{1-\alpha_t} (M^{\textrm{f}}_{t-L: t-1}), X_{t-L: t-1}\big)\Big] \nonumber \\
    &\leq \frac{1}{T}\sum_{t=1}^T \mathbb{M}\big[\mathcal{H}(M^{\textrm{f}}_{t-L: t-1}, X_{t-L: t-1})\big] + R \cdot \frac{1}{T}\sum_{t=1}^T \mathbb{M} \left[\left|Q_{1-\alpha_t} \left(M^{\textrm{f}}_{t-L: t-1}\right) - M^{\textrm{f}}_{t-L: t-1}\right|^{\beta}\right].
\end{align}

By comparing \eqref{eq_hold2} with \eqref{eq_expand2}, we can deduce the inequalities
\begin{align}\label{eq_mid3}
    \frac{1}{T}&\sum_{t=1}^T \mathbb{M}\Big[\mathcal{H}\big(Q_{1-\alpha_t}(M^{\textrm{f}}_{t-L: t-1}),X_{t-L: t-1}\big)\Big] \nonumber\\
    &< \frac{1}{T}\sum_{t=1}^T Q_{1-\alpha_t}\big(\mathcal{H}(M^{\textrm{f}}_{t-L: t-1},X_{t-L: t-1})\big) -\epsilon - 2\max\left\{R,1\right\}\left(\frac{C}{\sqrt{L}}\right)^{\min\left\{\beta,1\right\}} \nonumber\\
    &<\frac{1}{T} \sum_{t=1}^T Q_{1-\alpha_t}(M_{t-L,t-1}) - 2\max\left\{R,1\right\}\left(\frac{C}{\sqrt{L}}\right)^{\min\left\{\beta,1\right\}}
\end{align}
where the second inequality uses the length preservation assumption in \eqref{eq_len_pre}.

By using quantile stability assumption in \eqref{eq_quant_stab}, we have
\begin{align}
        \frac{1}{T} \sum_{t=1}^T \mathcal{H}\big(Q_{1-\alpha_t} (M^{\text{\rm{f}}}_{t-L: t-1}), X_t\big) & < \frac{1}{T}\sum_{t=1}^T \mathbb{M}\Big[\mathcal{H}\big(Q_{1-\alpha_t}(M^{\textrm{f}}_{t-L: t-1}),X_{t-L: t-1}\big)\Big] + \frac{C}{\sqrt{L}}\nonumber\\
        & < \frac{1}{T}\sum_{t=1}^T Q_{1-\alpha_t}\left(M_{t-L: t-1}\right) - 2\max\left\{R,1\right\}\left(\frac{C}{\sqrt{L}}\right)^{\min\left\{\beta,1\right\}} + \frac{C}{\sqrt{L}} \nonumber\\
        & < \frac{1}{T}\sum_{t=1}^T Q_{1-\alpha_t}\left(M_{t-L: t-1}\right),
    \end{align}
where the second inequality uses \eqref{eq_mid3}.

Setting the weights $\left\{w_t^{\tau}\right\}_{\tau=1}^{L+1}$ to be uniform reduces AOCP and AFOCP to OCP and FOCP, respectively. Therefore, by the same reasoning, FOCP yields a shorter time-averaged prediction interval length than OCP under the same long-term coverage target.

\bibliographystyle{IEEEtran}
\bibliography{cite.bib}
\end{document}